\documentclass[journal,comsoc]{IEEEtran}
\usepackage{amsmath,graphicx}
\usepackage{booktabs}
\usepackage{verbatim}
\usepackage{color}
\usepackage{amsfonts}
\usepackage{multirow}
\usepackage{setspace}
\usepackage{amssymb}
\usepackage{pifont}
\usepackage{pgfplots}
\usepackage{tikz}
\usepackage{subfigure}
\usepackage{caption}
\usepackage{threeparttable}
\usepackage[T1]{fontenc}% optional T1 font encoding
\usepackage[numbers,sort&compress]{natbib}
\usepackage{url}
\usepackage{verbatim}

\ifCLASSINFOpdf
\else
\fi
\usepackage{amsmath}
\usepackage[cmintegrals]{newtxmath}
\begin{document}
\bibliographystyle{unsrt} % For reference sorting
\title{All You Need is a Second Look: Towards Arbitrary-Shaped Text Detection}
%\title{NASK: Promoting Arbitrary-Shaped Text Detection with a two-stage segmentation architecture}
% All You Need Is Boundary: Toward Arbitrary-Shaped Text Spotting 

\author{Meng~Cao, %~\IEEEmembership{Member,~IEEE,}
	Can~Zhang, %~\IEEEmembership{Member,~IEEE,}
	Dongming~Yang,~\IEEEmembership{Member,~IEEE,}
        %John~Doe,~\IEEEmembership{Fellow,~OSA,}
        and~Yuexian~Zou$^{*}$,~\IEEEmembership{Senior Member,~IEEE}% <-this % stops a space
\thanks{M. Cao, C. Zhang, DM.Yang and Y. Zou are with the School of Electrical and Computer Engineering, Shenzhen Graduate School, Peking University, Shenzhen 518055, China (E-mail: mengcao, zhangcan, yangdongming, zouyx@pku.edu.cn;).}% <-this % stops a space
\thanks{Y. Zou is also with Peng Cheng Laboratory, Shenzhen, China.}
\thanks{Corresponding Author$^{*}$ : Y. Zou}% <-this % stops a space
}


\markboth{Journal of \LaTeX\ Class Files,~Vol.~14, No.~8, August~2015}%
{Shell \MakeLowercase{\textit{et al.}}: Bare Demo of IEEEtran.cls for IEEE Communications Society Journals}



\maketitle

\begin{abstract}
Arbitrary-shaped text detection is a challenging task since curved texts in the wild are of the complex geometric layouts. Existing mainstream methods follow the instance segmentation pipeline to obtain the text regions. However, arbitrary-shaped texts are difficult to be depicted through one single segmentation network because of the varying scales. In this paper, we propose a two-stage segmentation-based detector, termed as NASK (\textbf{N}eed \textbf{A} \textbf{S}econd loo\textbf{K}), for arbitrary-shaped text detection. Compared to the traditional single-stage segmentation network, our NASK conducts the detection in a coarse-to-fine manner with the first stage segmentation spotting the rectangle text proposals and the second one retrieving compact representations. Specifically, NASK is composed of a Text Instance Segmentation (TIS) network ($1^{st}$ stage), a Geometry-aware Text RoI Alignment (GeoAlign) module, and a Fiducial pOint eXpression (FOX) module  ($2^{nd}$ stage). Firstly, TIS extracts the augmented features with a novel Group Spatial and Channel Attention (GSCA) module and conducts instance segmentation to obtain rectangle proposals. Then, GeoAlign converts these rectangles into the fixed size and encodes RoI-wise feature representation. Finally, FOX disintegrates the text instance into serval pivotal geometrical attributes to refine the detection results. Extensive experimental results on three public benchmarks including Total-Text, SCUT-CTW1500, and ICDAR 2015 verify that our NASK outperforms recent state-of-the-art methods.

\end{abstract}


% \fi
%
% For peerreview papers, this IEEEtran comand inserts a page break and
\IEEEpeerreviewmaketitle

\section{Introduction}
\IEEEPARstart{S}{cene} text detection (STD) aims to accurately localize text regions given a natural scene image, and has attracted a surge of attention in computer vision community due to its practical applications. %The performance of STD has been greatly enhanced by the advanced object detection~\cite{girshick2015fast,liu2016ssd} and segmentation~\cite{long2015fully} frameworks. 
However, despite the significant achievements in multi-oriented text detection, arbitrary-shaped text detection still remains challenging due to the complex geometric layouts.

Multi-oriented text instances take the rectangle or quadrilateral bounding-box to represent the detection results~\cite{liao2016textboxes,liao2018textboxes++,tian2016detecting}. These simple representations, however, fall short when dealing with the more laborious arbitrary-shaped texts. Therefore, several segmentation-based methods~\cite{feng2019textdragon, long2018textsnake, liu2020abcnet} have been proposed to deal with such challenging yet universal scenario. Most of the current overwhelming majority of arbitrary-shaped text detection methods can roughly be classfied into two categories: top-down, global modeling methods~\cite{liu2020abcnet,wang2020textray} and bottom-up, local modeling methods~\cite{long2018textsnake, feng2019textdragon}. Typically, global modeling methods treat texts as a special type of object and directly take the regression-based methodology to obtain the detection results. For simplification, they often presuppose the distribution of the text instances (\textit{e.g.} Bezier assumption for ABCNet~\cite{liu2020abcnet} and Chebyshev polynomial approximation for TextRay~\cite{wang2020textray}), and reconstruct texts with regressed key points. Although these methods achieve faster inference speed, their pre-defined distribution hypothesises are empirical without universality, leading to the inferior performance. On the contrast, the bottom-up methods~\cite{long2018textsnake, feng2019textdragon} are more well-defined, \textit{i.e.,} they use several crucial geometry attributes to rebuild the whole text instances. For example, the pioneering work TextSnake~\cite{long2018textsnake} represents text instances with a set of serially-connected disks and achieves competitive detection results. However, it suffers from two limitations. 

\begin{figure}[t]
	\centering
	%\centerline{\includegraphics[width=8cm]{experiments/ablation/vsTextSnake.pdf}}
	\centerline{\includegraphics[width=0.8\linewidth]{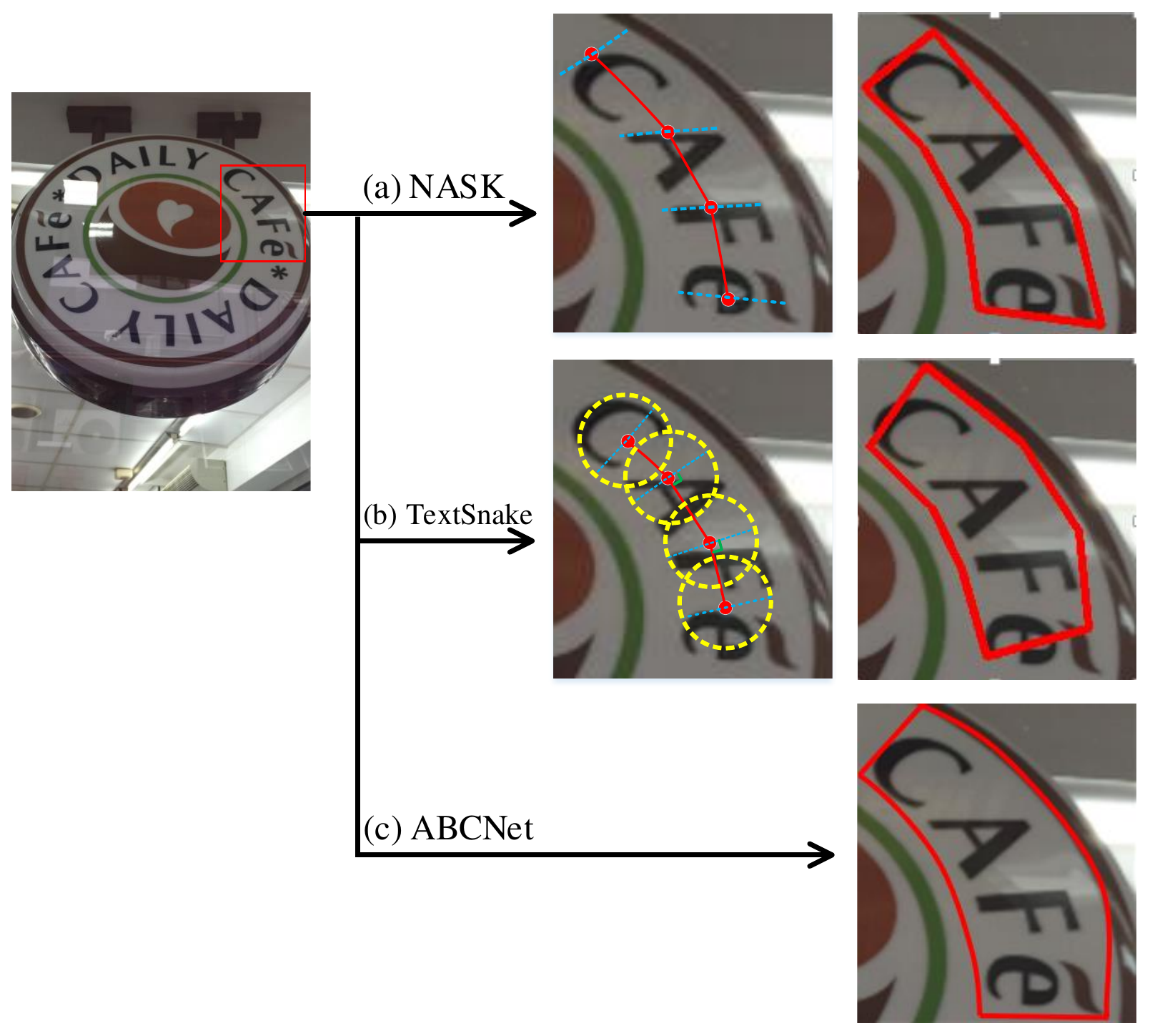}}
	%\centerline{(a) vsTextSnake}\medskip
	%\caption{Fiducial point generation result using the proposed method (Up) and the detection result of TextSnake (Down).}
	\caption{Typical curved text instance representations. (a) Our proposed NASK; (b) TextSnake; (c) ABCNet. NASK achieves a more compact and accurate representation than the others. TextSnake suffers from boundary character cut off while ABCNet generates a loose result with more background.}
	\label{fig:foxCompare}
\end{figure}   

Firstly, TextSnake applies the geometric attribute segmentation network on the input images directly, which makes it less resistant to noise~\cite{wang2020textray}. In some cases where the text instances are extremely tiny, the geometric attribute prediction becomes more difficult because even trivial segmentation deviation may lead to the ultimate failure. Therefore, a single segmentation network fails to process text instances that vary greatly in scales.

Secondly, the text geometric modeling in TextSnake is not optimal. As illustrated in Fig.~\ref{fig:foxCompare}(b), the control unit of each overlapping disk includes the text center line, disk radius, and the text line orientation. Implicitly, TextSnake regards the character direction to be perpendicular to the text direction. This assumption, however, is often too restrictive and may lead to failure detection especially when encountering distorted texts.
 %which can be divided into two categories: 1) Segmentation-based methods~\cite{zhang2016multi,yao2016scene}. These methods draw inspiration from instance segmentation and conduct dense predictions in the pixel level. 2) Regression-based methods~\cite{liao2018textboxes++,zhou2017east}. Scene texts are detected using the adapted one-stage or two-stage frameworks which have been proved effective in general object detection tasks.

%However, STD remains a challenging task due to its unique characteristics. Firstly, since the convolutional operation which is widely used in all segmentation or detection networks only processes a \textit{local} neighbour, it hardly captures the long-range dependencies even it is stacked repeatedly. Thus, CNN-based STD methods sometimes fail to detect long text instances because they are far beyond CNN's receptive field~\cite{wang2018non}. Secondly, although detecting words or text lines with a relatively simple rectangle or quadrilateral representation has been well tackled,  curve text detection with a tighter representation is not well solved~\cite{long2018textsnake}. %Secondly, segmentation-based methods are susceptible to background interference. It seems that these methods prefer a proposed region with only one target object included.
%Finally, some text instances are extremely tiny which makes their precise shape description more difficult because even a little segmentation deviation may lead to the ultimate failure. Therefore, a single segmentation network fails to process images that vary greatly in text scales.

To address those above issues, we propose a two-stage segmentation-based network termed as NASK for accurate arbitrary-shaped text detection. NASK consists of a Text Instance Segmentation network (TIS),  a Geometry-aware Text RoI Alignment module (GeoAlign) and a Fiducial pOint eXpression module (FOX). The benefits of using the two cascaded segmentation networks are two-fold: 1) With the first stage segmentation to obtain the rectangle text proposals, the second stage FOX utilizes Region of Interest (RoI) features to predict the basic geometry attributes. Compared to applying FOX on the whole input images, it reduces the background interference greatly. 2) We utilize GeoAlign to transform varying-size text instances to fixed-size feature maps before feeding them to the second stage network FOX and in this way, we save the FOX network from suffering from varying-size text input. Namely, text instances with varying shapes in input images are represented by fixed size feature maps, which eases the network training.

Specifically, the first stage segmentation network TIS is designed to localize rectangular text instances with the proposed Group Spatial and Channel Attention module (GSCA). Compared to the traditional Non-local neural network~\cite{wang2018non}, GSCA takes one step further to model long-range dependencies more extensively by computing interactions between any two positions across both space and channels. %which enlarges the receptive field of the shared feature maps. 
Then the GeoAlign module transforms the varying-size rectangular RoIs into a fixed size. Compared to the traditional RoI Pooling~\cite{ren2015faster} and RoI alignment~\cite{he2017mask}, our GeoAlign adaptively selects the sampling points and avoids the background interference. Finally, FOX is a novel arbitrary-shaped text representation which resorts to a set of fiducial points which are calculated with several pre-defined geometry attributes to generate accurate outputs. As shown in Fig.~\ref{fig:foxCompare}, in comparison to state-of-the-art methods~\cite{long2018textsnake,liu2020abcnet}, our NASK achieves tighter and more flexible results, thus more suitable for the arbitrary-shaped text reconstruction. 

In a nutshell, the main contributions of this work are as follows: (1) A novel attention module termed as GSCA is proposed to explore both the spatial-wise and channel-wise correlations in a more extensive way for more informative feature refinements. (2) We propose a more resonable representation tailored for arbitrary-shaped text instances called Fiducial pOint eXpression module (FOX). (3) We introduce a geometry-aware sampling method, a.k.a GeoAlign, for accurate RoI feature alignment. (4) Based on the novel two-stage segmentation architecture, our detector NASK achieves state-of-the-art performance on both curved and multi-oriented text detection benchmarks. 

The preliminary work has been published in ICASSP 2020~\cite{cao2020all} and we have extended it in the following significant aspects:
\begin{itemize}
\item We improve the previous Text RoI Pooling module~\cite{cao2020all} to the Geometry-aware RoI alignment module (GeoAlign). Experiment results show that GeoAlign leads to more accurate detection results.
\item We revisit our Group Spatial and Channel Attention Module and redesign the Global Channel Attention branch with a squeeze-and-excitation module~\cite{hu2018squeeze}. Compared to the previous version, it brings about better performance with negligible overheads.
\item Besides the curved text datasets, more experiments are conducted on the multi-oriented scene text dataset, which demonstrates that our proposed NASK is a more general text detector with state-of-the-art performance.
\end{itemize}
\begin{comment}
\begin{figure}[htb]
 \centering
  %\centerline{\includegraphics[width=9cm]{PipelineHalfColumnTwoLayer.pdf}}
  \includegraphics[width=9cm]{teaser/teaser.pdf}
   % \centerline{\includegraphics{PipelineHalfColumnOneLayer.pdf}}
%  \centerline{(a) Architecture}\medskip
\caption{The pipeline of NASK. \textbf{Above:} some key modules. \(1^{st}seg\) and \(2^{nd}seg\) means the first and the second stage segmentation networks respectively. \textbf{Below:} The illustration of an image instance transformation process.}
\label{fig:res}
\end{figure}

\begin{figure*}[htb]
	\centering
	%\centerline{\includegraphics[width=9cm]{PipelineHalfColumnTwoLayer.pdf}}
	\includegraphics[width=1\linewidth]{teaser/pipeline_fullStage_v3.pdf}
	% \centerline{\includegraphics{PipelineHalfColumnOneLayer.pdf}}
	%  \centerline{(a) Architecture}\medskip
	\caption{The pipeline of NASK. \(1^{st}seg\) and \(2^{nd}seg\) means the first and the second stage segmentation networks, respectively. In GeoAlign, RoI-wise affine transformations are predicted and embedded in feature map $\mathcal{T}$ to generate the geometry-aware representation feature map $V$.}
	\label{fig:pipeline}
\end{figure*}
\end{comment}

\section{Related work}
In Section~\ref{subsec:std}, we review the recent progress of scene text detection. Specifically, we analyze the current curved text representation methods in Section~\ref{subsec:curveRepresent}. Finally, Section~\ref{subsec:selfAtten} inspects the self-attention mechanism and its variants.
\subsection{Scene Text Detection}\label{subsec:std}
Based on deep neural networks, scene text detection methods have progressed extensively. These methods are roughly classified into three categories.

\textbf{Component-based Methods:} This kind of methods detect text parts or components as well as the linking relationships between them and then group them into final results through post-processes. CTPN~\cite{tian2016detecting} detects the text line in a sequence of vertical anchors that jointly predict location and text/non-text scores. Seglink~\cite{shi2017detecting} decomposes text instances into two elements, namely segments and links, where a segment is an oriented box covering a part of a word while a link indicates whether two segments belong to the same word or not. WordSup~\cite{hu2017wordsup} proposes a weakly-supervised framework that can utilize word annotations in both tight quadrangles and the more loose bounding boxes. MCN~\cite{liu2018learning} generates text bounding boxes by firstly converting an image into a Stochastic Flow Graph (SFG) and then performing Markov Clustering on this graph.

\textbf{Detection-based Methods:} Scene texts are detected using the adapted one-stage or two-stage frameworks which have been proved effective in general object detection tasks. TextBoxes~\cite{liao2016textboxes} inherits the architecture of SSD~\cite{liu2016ssd} and makes some adaptive modifications to achieve both high accuracy and efficiency in a single network forward pass. TextBoxes++~\cite{liao2018textboxes++} extends TextBoxes to  handle multi-oriented texts and refines end-to-end text recognition combined with a text recognizer. RRPN~\cite{ma2018arbitrary} proposes a Rotation Region Proposal Network which generates inclined proposals with text orientation angle information to facilitate the multi-oriented text detection. EAST~\cite{zhou2017east} is another one-stage based detector which directly predicts text instances with arbitrary orientations and quadrilateral shapes in full images. Liao \textit{et al.}~\cite{liao2018rotation} propose a rotation-sensitive regression for oriented scene text detection, which extracts the rotation-sensitive and rotation-invariant features for the regression and the classification branch respectively.

\textbf{Segmentation-based Methods:} Segmentation-based methods draw inspiration from instance segmentation and conduct dense predictions in the pixel level. Zhang \textit{et al.}~\cite{zhang2016multi} detect multi-oriented scene text using Fully Convolutional Network (FCN) model to predict the salient map of text regions in a holistic manner. Lyu \textit{et al.}~\cite{lyu2018multi} propose to detect scene texts by localizing corner points of text bounding boxes and segmenting text regions in relative positions. PSENet~\cite{wang2019shape} applies different scales of kernels for each text instance and generates the corresponding scale segmentation maps. Based on Mask R-CNN, SPCNet~\cite{xie2019scene} proposes a supervised pyramid context network to suppress false positives and achieves better performance when applied to rotated texts.

\textbf{Discussion:} Though detection-based methods tend to have the competitive performance on quadrilateral text detection, many of them fail to deal with curved texts limited by their baseline algorithms. In contrast, segmentation-based methods can naturally handle the more general arbitrary-shaped text case. However, they are more subject to background interference and more sensitive to the segmentation deviation~\cite{wang2020textray}. To alleviate this dilemma, we design a two-stage segmentation network with the first stage to locate the rectangular text instances and the second stage to reconstruct a compact representation. Experimental results show that our two-stage architecture is more robust and efficient.

\subsection{Arbitrary-Shaped Text Representation}\label{subsec:curveRepresent}
Arbitrary-shaped text instances are of irregular layout and the conventional representations such as axis-aligned rectangles or quadrangles struggle with giving precise modeling. Several representations are proposed to fit texts of arbitrary shapes. TextSnake~\cite{long2018textsnake} describes text instances with a series of ordered, overlapping disks, each of which is sampled along the text center line and associated with specific radius and orientations. The final text shape is composed of circular circumscribed polygons. In this representation, the line between the tangent point and the corresponding center of the circle is perpendicular to the centerline, which may not be the most reasonable case. %Implicitly, TextSnake regards the character direction to be perpendicular to the text direction by default, which may lead to failure detection especially when encountering distorted texts. 
In NASK, we apply an additional character orientation prediction which models the text character direction. As illustrated in Fig.~\ref{fig:foxCompare}, with the added character orientation prediction, the proposed method has a more accurate and flexible representation, resulting in better detection results.

ABCNet~\cite{liu2020abcnet} adopts a parameterized Bezier curve to fit arbitrarily-shaped texts. Specifically, it simplifies the detection problem to the control point regression problem, based on which the Bezier curve is generated. However, the Bezier curve assumption based on sparse control points is too restrictive and the Bernstein Polynomials~\cite{lorentz2013bernstein} may not be the optimal solution. Besides, empirically, it tends to generate loose regions partially because of the sparse control points and fails to output compact detection results as shown in Fig.~\ref{fig:foxCompare} (c). Compared to ABCNet, we do not presuppose the composition of the curve but represent it by a set of boundary points, which is more flexible and tighter.

%\textbf{Specific comparisons with TextSnake.}
%TextSnake shares the similar text construction strategy with our proposed method which treats a text instance as a sequence of overlapping disks and each disk is represented by a set of geometric properties including text center line, disk radius and the text line orientation. Implicitly, TextSnake regards the character direction to be perpendicular to the text direction by default, which may leads to the failure detection especially when encountering distorted texts. For fair comparison, the two-stage architecture is also implemented in TextSnake. As illustrated in Fig.~\ref{fig:foxCompare}, with the additional character orientation prediction, the proposed method has a more accurate and flexible representation, resulting in a better detection result.

\subsection{Self-attention}\label{subsec:selfAtten}
The self-attention mechanism has been proved to be effective in machine translation task~\cite{vaswani2017attention}. Besides, it has also been widely used in other areas such as computer vision.
%since the convolutional operation which is widely used in all segmentation or detection networks only processes a \textit{local} neighbour, it hardly captures the long-range dependencies even it is stacked repeatedly. Thus, CNN-based STD methods sometimes fail to detect long text instances because they are far beyond CNN's receptive field~\cite{wang2018non}.

\textbf{Non-local Operations:}
\cite{wang2018non} proposes the non-local neural network based on the self-attention mechanism which computes the response at a position as a weighted sum of the features within all same-channel positions. Due to its superior performance, it has been widely used in object detection and segmentation. \cite{yuan2018ocnet} applies the adapted non-local modules to increasing the resolution of feature maps in a coarse-to-fine manner, resulting in more accurate results. %Double Attention Network~\cite{fu2019dual} proposes a two-stage attention network with the first one gathering features from the entire space into a compact set and the second one adaptively distributing features to every location. 
DANet~\cite{fu2019dual} appends two types of attention modules, which model the semantic interdependencies in spatial and channel dimensions respectively.
CCNet~\cite{huang2019ccnet} captures the lone-range dependency information in a more efficient way, namely only considering the correlations among pixels on the criss-cross path.

\textbf{Channel Correlations:}
SENet~\cite{hu2018squeeze} is probably the first structure to explicitly model the interdependencies between channels and adaptively recalibrates channel-wise feature responses. SCA-CNN~\cite{chen2017sca} incorporates both spatial and channel-wise attention in a CNN to facilitate the task of image caption. ~\cite{lu2018channel} proposes a compact channel attention combined with the multi-level feature fusion mechanism which benefits image super-resolution.

\textbf{Discussion:} Different from previous works, our GSCA extends the self-attention mechanism in both spatial and channel perspectives, and carefully design the spatial and global channel attention branches to capture rich contextual relationships for better feature representations. The primary work DANet~\cite{fu2019dual} also adopts a dual attention mechanism from both the spatial and channel perspectives. Our work, however, differs in the following two respects. Firstly, in the spatial attention module, GSCA takes a more radical approach that computes the correlations among all the elements in the feature map, not limited in the same-channel interrelationship. Secondly, in the channel attention branch, instead of computing the channel-wise correlations in DANet, we use a more simple yet efficient Squeeze-Excitation-like module~\cite{hu2018squeeze} to achieve better performance while preserving the efficiency.

\section{Approach}
\begin{figure*}[htb]
	\centering
	%\centerline{\includegraphics[width=9cm]{PipelineHalfColumnTwoLayer.pdf}}
	\includegraphics[width=1\linewidth]{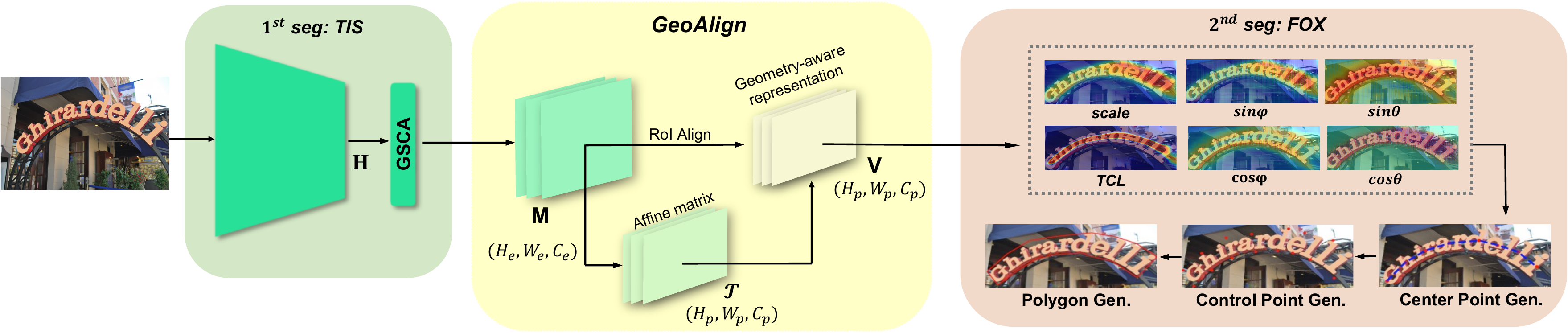}
	% \centerline{\includegraphics{PipelineHalfColumnOneLayer.pdf}}
	%  \centerline{(a) Architecture}\medskip
	\caption{The pipeline of NASK. \(1^{st}seg\) and \(2^{nd}seg\) means the first and the second stage segmentation networks, respectively. In GeoAlign, RoI-wise affine transformations are predicted and embedded in feature map $\mathcal{T}$ to generate the geometry-aware representation feature map $V$.}
	\label{fig:pipeline}
\end{figure*}
The proposed NASK achieves accurate text representation and is more robust facing varying-size input. In this section, we first present an overview of the whole pipeline (Section~\ref{subsec:overview}). Then we investigate three proposed modules including GSCA (Section~\ref{subsec:gsca}), GeoAlign (Section~\ref{subsec:geoPool}) and FOX (Section~\ref{subsec:FOX}), respectively. Finally, the optimization details are presented in Section~\ref{subsec:optim}.

\subsection{Overview}\label{subsec:overview}
The overall pipeline of NASK is presented in Fig~\ref{fig:pipeline}. It consists of three cascaded components: the first stage segmentation network TIS, a geometry-aware RoI transformation module GeoAlign and the second stage segmentation network FOX. 

An input image is passed through the first stage segmentation network TIS, which is designed in a fully convolutional fashion~\cite{long2015fully}. In order to efficiently aggerate the contextual information of the generated feature map $\mathbf{H}$, we append a Group Spatial and Channel Attention Module after the fully convolutional network to obtain the more informative feature map $\mathbf{M}$.

Given the refined feature map $\mathbf{M}$, we first threshold on pixel to obtain the binary classcification map, \textit{i.e.} text or non-text areas respectively. Methods like \textit{minAreaRect} in OpenCV~\cite{bradski2008learning} is applied to group the predicted positive pixels into rectangle Connected Components. Then for the convenience of the next stage input, the cropped feature maps are required to be transformed into a fix size. Thus, we apply GeoAlign to conduct RoI-wise transformation. With GeoAlign, in addition to achieving the desired size normalization, we also obtain a more geometry-aware RoI feature representation, shown as the feature map $\mathbf{V}$ in Fig.~\ref{fig:pipeline}.

Then, we feed the RoI features into a relatively simple segmentation network FOX with several convolution and up-sampling layers. The final output layer of FOX contains 6 channels, which represent the prediction of geometry attributes including text center line (TCL), character scale, character orientation and text orientation. Finally, text polygons are generated by applying \textit{approxPolyDP} in OpenCV based on the detected fiducial points.

\subsection{Group Spatial and Channel Attention Module}\label{subsec:gsca}
The conventional convolution is inherently a regional operation and limited to local receptive fields. The generated feature map with insufficient contextual information imposes a great adverse effect on the downstream tasks. To model comprehensive dependencies over local feature representations, we introduce a Group Spatial and Channel Attention Module, GSCA for short. GSCA captures contextual information in both spatial and channel aspects. In spatial relationship modeling, we explicitly learns the correlations among all elements of the whole feature map. Compared to Non-local Neural Network~\cite{wang2018non} which constrains the correlation modeling within the same channel, GSCA exploits the spatial dependencies in a more radical way. In order to alleviate the huge computational overhead, we introduce the \textit{channel grouping idea} to split all \textit{C} channels into \textit{G} groups and only the intra-group relationships (each group with $C' = C/G$ channels) are estimated. To capture the inter-group correlations, a Global Channel Attention branch is devised to generate the channel-wise attention and distribute information among every group. %In this way, we have mitigated the disadvantages of grouping 

\begin{figure}[htb]
 \centering
 % \centerline{\includegraphics[width=9cm]{approach/selfAttenattention_jrnl.pdf}}
\includegraphics[width=1\linewidth]{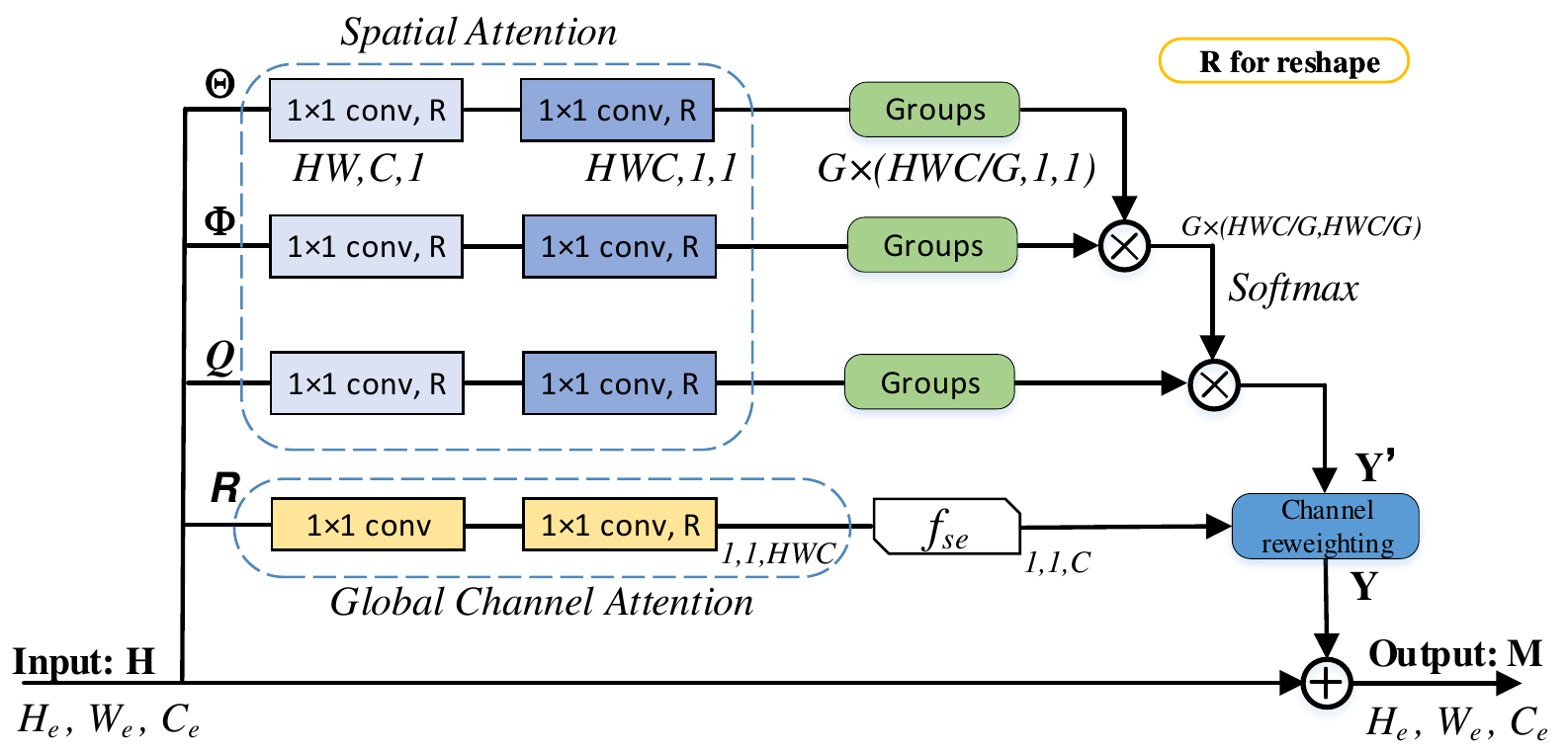}
  %\centerline{(a) Attention}\medskip
\caption{Group Spatial and Channel Attention module: Intra-group attention is learned by the serially connected spatial \textit{convolution} and \textit{reshape} denoted as \textit{\(\Theta\)}, \(\Phi\), $Q$ while the global channel attention is captured by transformation \textbf{R} and $f_{se}$. "\(\oplus\)" denotes the element-wise sum while "\(\otimes\)" denotes matrix multiplication. The annotations under each block represent the corresponding output size.}
\label{fig:gscaPipline}
\end{figure}

As shown in Fig.~\ref{fig:gscaPipline}, given the backbone generated feature map $\mathbf{H} \in \mathbb{R}^{H_e \times W_e \times C_e}$, for each position $\mathbf{u}$ in $\mathbf{H}$, we generate the intra-group affinity map $\mathbf{A} \in \mathbb{R}^{(H_eW_eC_e/G) \times(H_eW_eC_e/G)}$. Specifically, the spatial-attended feature map $\mathbf{Y'}$ is generated as follows.

\begin{equation}
\mathbf{Y'} = concat\Big(f\big(g(\Theta(\mathbf{H})), g(\Phi(\mathbf{H}))\big)g(Q(\mathbf{H}))\Big), \label{con:intraAtten}
\end{equation}

\noindent where %$\boldsymbol{H} \in \mathbb{R}^{H_e \times W_e \times C_e}$ denotes the input feature map. 
$\Theta(\mathbf{H})$, $\Phi(\mathbf{H})$, $Q(\mathbf{H})$ are learnable spatial transformations implemented as serially connected \textit{convolution} and \textit{reshape}. $f(\cdot, \cdot)$ is defined as matrix product for simplification and $g$ is the grouping operation which divides the feature map into $G$ groups along the channel dimension. $concat$ denotes the channel-wise concatenation. Therefore, the output $\mathbf{Y'}$ shares the same shape as input $\mathbf{H}$. 

%Then we have \(Y' = \Theta(X)\Phi(X)^Tg(X)\) where \(Y'\) is the \textit{group} result of \(Y\).
As for the Global Channel Attention branch, we capture the channel-wise weights $\lambda$ with a squeeze-and-excitation module:

\begin{equation}
\boldsymbol{\lambda} = f_{se}(R(\mathbf{H}))= softmax\Big(W_2\big(\sigma (W_1 H(\mathbf{H}))\big)\Big),
\end{equation}

\noindent where $\boldsymbol{\lambda} \in \mathbb{R}^{1 \times 1 \times C}$ is the chanel-wise attention weight and $f_{se}$ is the excitation function. Specifically, $\sigma$ denotes the ReLU function, $W_1 \in  \mathbb{R}^{\kappa T \times T}$  and $W_2 \in  \mathbb{R}^{T \times \kappa T}$ ($\kappa$ is the expansion ratio) are the learnable parameters of two fully-connected layers. Thus, we apply the channel-wise reweighting as follows. 

\begin{equation}
\mathbf{Y} = \lambda_i \mathbf{Y'_i},
\end{equation}

\noindent where $i \in [1, C]$ is the channel index and $C$ is the number of channels. $\lambda_i$ and $\mathbf{Y'_i}$ denote the $i$-th channel weight and $i$-th channel feature map respectively. Meanwhile, a short-cut path is used to preserve the local information and the final output $\mathbf{M}$ is the sum of $\mathbf{H}$ and $\mathbf{Y}$. 

\begin{equation}
\mathbf{M} = \mathbf{H} + \mathbf{Y}.
\end{equation}

\subsection{Geometry-aware RoI Alignment Module}\label{subsec:geoPool}
\begin{comment}
\begin{figure}[htb]
	\centering
	\centerline{\includegraphics[width=7.5cm]{approach/stage2.pdf}}
	%\centerline{(a) Attention}\medskip
	\caption{Stage2}
	\label{fig:StagePool}
\end{figure}
\end{comment}

\begin{figure}[htb]
	\centering
	\subfigure[]{
	\begin{minipage}[b]{0.2\linewidth}
		\centerline{\includegraphics[width=6cm]{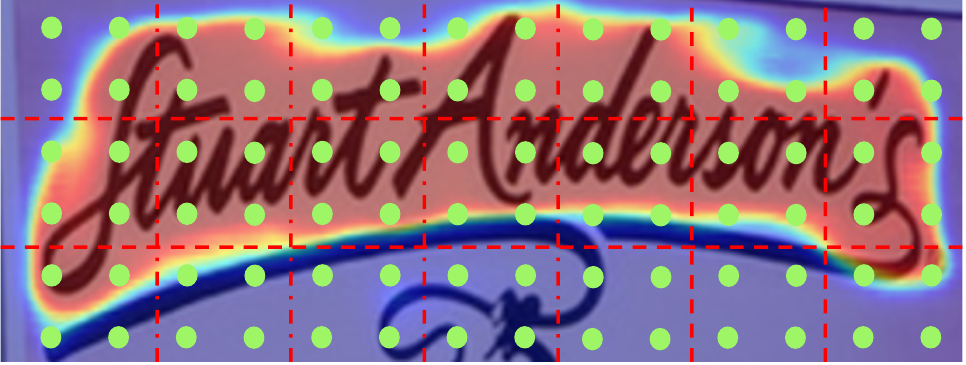}}
	\end{minipage}
	}	

	\subfigure[]{
	\begin{minipage}[b]{0.2\linewidth}
		\centerline{\includegraphics[width=6cm]{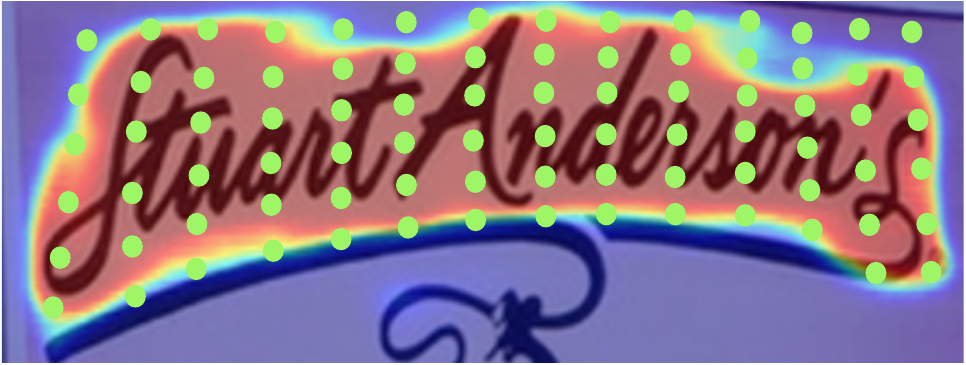}}
	\end{minipage}
}
	%\centerline{(a) Attention}\medskip
	\caption{(a) RoI Align uniformly applies the pooling procedure with $k^2$ sample points in each bin, which brings about the background interference. (b) Our Geometry-aware RoI Alignment module adaptively selects the sample points within the text instances.} 
	\label{fig:GeoRoIAlign}
\end{figure}
To transform the varying-size RoIs into the fixed size, we have to apply a pooling-like module. In our previous work~\cite{cao2020all}, we simply apply the RoI Pooling module~\cite{ren2015faster} to obtain the cropped RoI feature maps. ~\cite{he2017mask} has demonstrated that RoI Align is a better substitute for RoI Pooling. As shown in Fig.~\ref{fig:GeoRoIAlign}(a), RoI Align sets the sampling points in a uniform way, namely it averages $k^2$ points in each bin and then applies max-pooling. Due to the characteristic of curved texts, some sampling points are outside the text areas, which inevitably brings about the background interference. To address this problem, we take one step further to develop the Geometry-aware RoI Alignment module which adaptively samples points within the text areas. Before we specify our GeoAlign, let’s revisit the RoI Align in detail.

For RoI Align, mathematically, given the RoI feature map $\mathbf{M} \in \mathbb{R}^{H_e\times W_e\times C_e}$ generated by the backbone network with GSCA, we have the following pooling feature map $\mathbf{V}\in \mathbb{R}^{H_p\times W_p\times C_p}$:
%After the carefully designed backbone network, we apply two sets of convolutions.

\begin{equation}
\boldsymbol{\mathbf{V}}_{i j}=\frac{1}{k^{2}} \sum_{x=k i}^{k(i+1)-1} \sum_{y=k j}^{k(j+1)-1} \mathbf{M}(p(x, y)),
\end{equation}

\noindent where $i \in[1, W_p]$, $j \in[1, H_p]$ denote the pixel index and $k^2$ is number of sampling points within each bin. $(x, y)$, $x \in [1, kW_p]$, $y \in [1, kH_p]$ is the horizontal and vertical sampling point index and $p(x, y)$ is the corresponding spatial position.

For GeoAlign, it adopts an additional affine transformation matrix $\mathcal{T}_{i j}$ to encode the geometry characteristics of the text information (e.g. rotation, translation, scale, and shear) and warps the uniformly sampling points to get the geometry-aware representation $\text{V}$ as follows.

\begin{equation}
\mathbf{V}_{i j}=\frac{1}{k^{2}} \sum_{x=k i}^{k(i+1)-1} \sum_{y=k j}^{k(j+1)-1} \mathbf{M} (\mathbf{\mathcal{T}}(p(x, y))), \label{con:calGeoPool}
\end{equation}

\noindent where $\mathbf{\mathcal{T}} \in  \mathbb{R}^{kH_p\times kW_p\times 6}$ is the warping parameters for each sampling point. Specifically, the affine warping transformation process is as follows.
\begin{equation}
\mathcal{T}(p(x, y))=\left[\begin{array}{lll}
\mathcal{T}_{x, y, 1} & \mathcal{T}_{x, y, 2} & \mathcal{T}_{x, y, 3} \\
\mathcal{T}_{x, y, 4} & \mathcal{T}_{x, y, 5} & \mathcal{T}_{x, y, 6}
\end{array}\right]\left(\begin{array}{l}
p(x) \\
p(y) \\
\;\;1
\end{array}\right),
\end{equation}

\noindent where $\mathcal{T}_{x, y, i}$, $i \in [1, 6]$ represents 6-dimensional affine parameters for each sampling point position $p(x, y)$. $p(x)$ and $p(y)$ are the horizontal and vertical components of $p(x, y)$, respectively.

\textbf{\textit{How to supervise the warping process?}} Namely, how to obtain the ground truth of the warped sampling points? Here, we take advantage of the boundary point annotations in dataset and use the bilinear interpolation to obtain more dense sampling points. Finally, a simple L1 loss for the wrapped points is adopted and the details are presented in Equation~\ref{con:lossPool}.

%\textbf{\textit{How to obtain the ground truth of warpped points?}}

\subsection{Fiducial point expression module}\label{subsec:FOX}
\begin{figure}[htb]
 \centering
  \centerline{\includegraphics[width=1\linewidth]{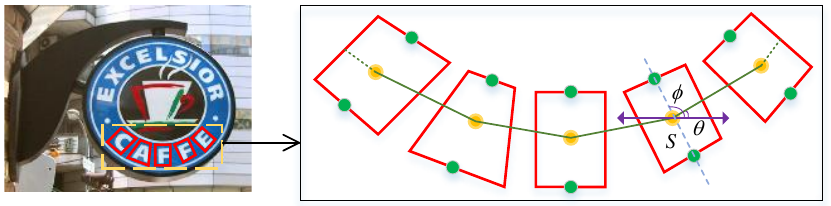}}
  %\centerline{\includegraphics{shapeRepresentation.pdf}}
  %\centerli ne{(a) shape representation}\medskip
\caption{Illustration of Fiducial Points Expression module. Center points are marked as yellow and fiducial points are marked as green.}
\label{fig:fox}
\end{figure}

Building an appropriate representation for arbitrary-shaped texts plays an important role in accurate detection. We leverage on the fiducial points of the text instances to build an accurate and flexible representation. The detail illustration of our FOX is depicted in Fig.~\ref{fig:fox}. The geometrical attributes utilized to make up the text instances include the text center line (TCL), the character scale $s$, the character orientation $\phi$  and the text orientation $\theta$. 

Mathematically, a text instance can be viewed as an ordered sequence $S = \{S_1, ..., S_i, ..., S_n\}$, where $n$ is the number of character segments. Each component $S_i$ is a free-form quadrilateral. We  construct the center point list $C = ( c_{start}, c_1, ..., c_i, ..., c_n, c_{end} )$, in which $c_i$ is the center point (marked as yellow in Fig.~\ref{fig:fox})  of $S_i$. Note that ${c}_{start}$ is the midpoint of $S_1$'s left edge and ${c}_{end}$ is the midpoint of $S_n$'s right edge. The center point list $C$ is evenly sampled from the text center line (a side-shrunk version of text polygon annotations following~\cite{long2018textsnake}).

Following the above notations, we define $S_i = (c_i, s_i, \phi_i, \theta_i)$. The fiducial points (marked as green in Fig.~\ref{fig:fox}) are defined as the midpoints of the top and bottom edges of each character quadrilateral. Thus, we compute the scale $s_i$ as half the height of the character while the character orientation $\phi_i$ is the direction from the bottom-edge midpoint to the corresponding top-edge one. For the text orientation $\theta_i$, it is defined as the horizontal angle between the current center $c_i$ and the next one $c_{i+1}$. 

%The scale $s_i$ is half the height of the character while the text orientation $\theta_i$ is defined as the horizontal angle between the current quadrilateral center $c_i$ and the next  one $c_{i+1}$. We take the midpoints on the top and bottom edges of each character quadrilateral as fiducial points and the character orientation $\phi_i$ is defined as the direction from the midpoint of the bottom edge to that of the top edge.

%Specifically, the text center line is a binary mask based on the side-shrunk version of text polygon annotations~\cite{long2018textsnake}. The scale $s_i$ is half the height of the character while the text orientation $\theta_i$ is defined as the horizontal angle between the current quadrilateral center $c_i$ and the next  one $c_{i+1}$. We take the midpoints on the top and bottom edges of each character quadrilateral as fiducial points and the character orientation $\phi_i$ is defined as the direction from the midpoint of the bottom edge to that of the top edge.

%Mathematically, a text instance can be viewed as an ordered sequence $S = \{S_1, ..., S_i, ..., S_n\}$, where $n$ is a hyper-parameter which denotes the number of character segments. Each node $S_i$ is associated with a group of geometrical attributes and can be represented as $S_i = (c_i, s_i, \phi_i, \theta_i)$, where every element is defined as above.

Based on the delicated designed fiducial point expression module, we set up a relatively simple segmentation network to generate the text polygon. The whole procedure can be divided into the following three steps.

\textbf{Center point Generation.}
Firstly, two up-sampling layers followed by one $1\times1$ convolution layer make up the full second stage segmentation network. Note that the final convolution layer is with 6 output channels to regress all the above geometrical attributes. Formally, the output is $F=\{f_1,f_2,...,f_6\}$ where $f_1$, $f_2$ denote the pixel-wise character scale $s$  and the probability belonging to TCL respectively. After thresholding the feature map $f_2$, we obtain the text center line areas. Then center point list $C = ( c_{start}, c_1, ..., c_i, ..., c_n, c_{end} )$ are equidistantly sampled along the center line. 

\textbf{Fiducial Point Generation.}
We use $f_3$ and $f_4$ to model the text orientation $\theta$ via its sine and cosine value. $sin \theta$ and $cos \theta$ are normalized to ensure their quadratic sum equals to 1:

\begin{equation} \label{eqn2}
\begin{split}
\cos \theta&=\frac{f_{3}}{\sqrt{f_{3}^{2}+f_{4}^{2}}} \\
\sin \theta&=\frac{f_{4}}{\sqrt{f_{3}^{2}+f_{4}^{2}}}.
\end{split}
\end{equation}

\noindent $sin \phi$ and $cos \phi$ are normalized with $f_5$ and $f_6$ in the same way. 

For each $c_i$ which has beed obtained in the preceding step, according to the geometric relationship, two corresponding fiducial points in the bottom and top edges are computed as follows.
%Given these geometrical attributes, we can acquire the boundary fiducial points using the following procedure. Firstly, \(k\) points are equidistantly sampled for the center line \(C\), named \( \overline{C} = ( \overline{c}_1, ..., \overline{c}_i, ..., \overline{c}_k ) \). For each \(\overline{c}_i\), the two corresponding fiducial points are computed as follows.

\begin{equation} \label{eqn2}
  \begin{split}
   p_{2i-1} &= {c}_i + (s_i cos\phi_i, -s_i sin\phi_i)    \\
   p_{2i} &= {c}_i + (-s_i cos\phi_i, s_i sin\phi_i),
  \end{split}
\end{equation}

\noindent where $p_{2i-1}$ is the top-edge fiducial point for the center point $c_i$ while $p_{2i}$ is its bottom-edge counterpart. $s_i$ and $\phi_i$ are the scale and the orientation for the $i$-th character respectively. Therefore, each text instance can be represented with $2n$ fiducial points. 

\textbf{Text Polygon Generation.}
Based on the obtained $2n$ fiducial points., we generate the text polygon for each instance via \textit{approxPolyDP} in OpenCV~\cite{bradski2008learning} which approximates the polygon with given vertices.

\begin{comment}
\begin{figure*}[htb]
 \centering
  %\centerline{\includegraphics[width=8.5cm]{controlPointGeneration.pdf}}
    \centerline{\includegraphics[width=10cm]{approach/controlPointGeneration_crossColumn.pdf}}
  %\centerline{(a) Architecture}\medskip
\caption{The illustration of Fiducial Point Expression. Note that, \#c stands for the number of channels.}
\label{fig:res}
\end{figure*}
\end{comment}
\subsection{Optimization}\label{subsec:optim}
The overall loss function contains three terms corresponding to the three modules:

\begin{equation}
    \mathcal{L} =  \mathcal{L}_{\text{TIS}} + \alpha \mathcal{L}_{\text{Align}} + \beta \mathcal{L}_{\text{FOX}},
\end{equation}

%For an input image \(I\), the loss is comprised of two parts as shown in Eq (2).  \(L_{stage1}\) is the loss for the first text instance region segmentation.
\noindent where \(\mathcal{L}_{\text{TIS}}\), \(\mathcal{L}_{\text{Align}}\) and $\mathcal{L}_{\text{FOX}}$ are the loss for Text Instance Segmentation, the Geometry-aware RoI Alignment module and the Fiducial Point Expression module, respectively. 

\(\mathcal{L}_{\text{TIS}}\) is implemented as a cross-entropy loss with OHEM~\cite{shrivastava2016training} adopted:

\begin{equation}
\mathcal{L}_{\text{TIS}}=\frac{1}{H W N} \sum_{i=1}^{H} \sum_{j=1}^{W} \sum_{n=1}^{N}-\log \left(p_{n}\left(\mathbf{M}_{i, j}\right)\right),
\end{equation}

\noindent where $H$ and $W$ are the height and width of the output feature map $\mathbf{M}$ of the first stage segmentation network TIS. $N$ represents the number of classification categories and here we set $N = 2$ for text and non-text areas respectively. $p_{n}(\mathbf{M}_{i, j})$ is the softmax score for the \textit{n}-th class of the pixel $\mathbf{M}_{i, j}$.

To supervise the training for the Geometry-aware alignment module, we apply the L1 loss for the sampling point warping.

\begin{equation}
\mathcal{L}_{\text{Align}} =\frac{1}{H_pW_pk^{2}}\left|\mathcal{T}(p({x, y}))-  p^{*}\left (x, y\right)\right|, \label{con:lossPool}
\end{equation}

\noindent where $H_p$ and $W_p$ denote the shape of the output pooling feature map. $x \in [1, kW_p]$ and $y \in [1, kH_p]$ are the horizontal and vertical index of the sampling points ($k^2$ points within each bin). $p^{*}(x, y)$ is the ground truth position, which is calculated by the interpolation of boundary points.

$\mathcal{L}_{\text{FOX}}$ represents the loss for all the regressed geometry attributes:

\begin{equation}
  \begin{split}
    \mathcal{L}_{\text{FOX}} = &\lambda_1 \mathcal{L}_{tcl} + \lambda_2 \mathcal{L}_{s} + \lambda_3 \mathcal{L}_{sin\theta} \\
    &+ \lambda_4 \mathcal{L}_{cos \theta} + \lambda_5 \mathcal{L}_{sin\phi} + \lambda_6 \mathcal{L}_{cos\phi}
     %P = \left( {P_0, P_1, ..., P_i,..., P_m;} \right. \\ \left. { P_{m+1}, P_{m+2}, ..., P_{m+1+i}, ..., P_{2m+1}} \right)  
    \end{split},
\end{equation}

\noindent where $\mathcal{L}_{tcl}$ is the cross-entropy loss for TCL areas. \(\mathcal{L}_s\), \(\mathcal{L}_{sin\theta}\), \(\mathcal{L}_{cos\theta}\), \(\mathcal{L}_{sin\phi}\) and \(\mathcal{L}_{cos\phi}\) are the Smoothed-L1 loss~\cite{ren2015faster} as follows:

\begin{equation}
\left(
\begin{array}
{c}
\mathcal{L}_s\\
\mathcal{L}_{sin\theta}\\
\mathcal{L}_{cos\theta}\\
\mathcal{L}_{sin\phi}\\
\mathcal{L}_{cos\phi}
\end{array}
\right)=SmoothedL1{
\left( \begin{array}{c}
\frac{\widehat{s}-s}{s}\\
\widehat{sin\theta} - sin\theta\\
\widehat{cos\theta} - cos\theta\\
\widehat{sin\phi} - sin\phi\\
\widehat{cos\phi} - cos\phi
\end{array} 
\right )},
\label{con:smoothL1}
\end{equation}

\noindent where $\widehat{s}$, $\widehat{sin\theta}$, $\widehat{cos\theta}$, $\widehat{sin\phi}$, $\widehat{cos\phi}$ are the predicted values while $s$, $sin\theta$, $cos\theta$, $sin\phi$, $cos\phi$ are the corresponding ground truth. 

%All pixels outside the TCL are set to 0 since the geometrical attributes make no sense to non-TCL points. 
The hyper-parameters $\lambda_1$, $\lambda_2$, $\lambda_3$, $\lambda_4$, $\lambda_5$, $\lambda_6$, $\alpha$, $\beta$ are all set to 1 in our experiments.

\section{Experiments}
\begin{figure*}[htbp]
	\centering
	\subfigure[Total Text]{
		\includegraphics[width=0.2\textwidth,height=0.15\textwidth]{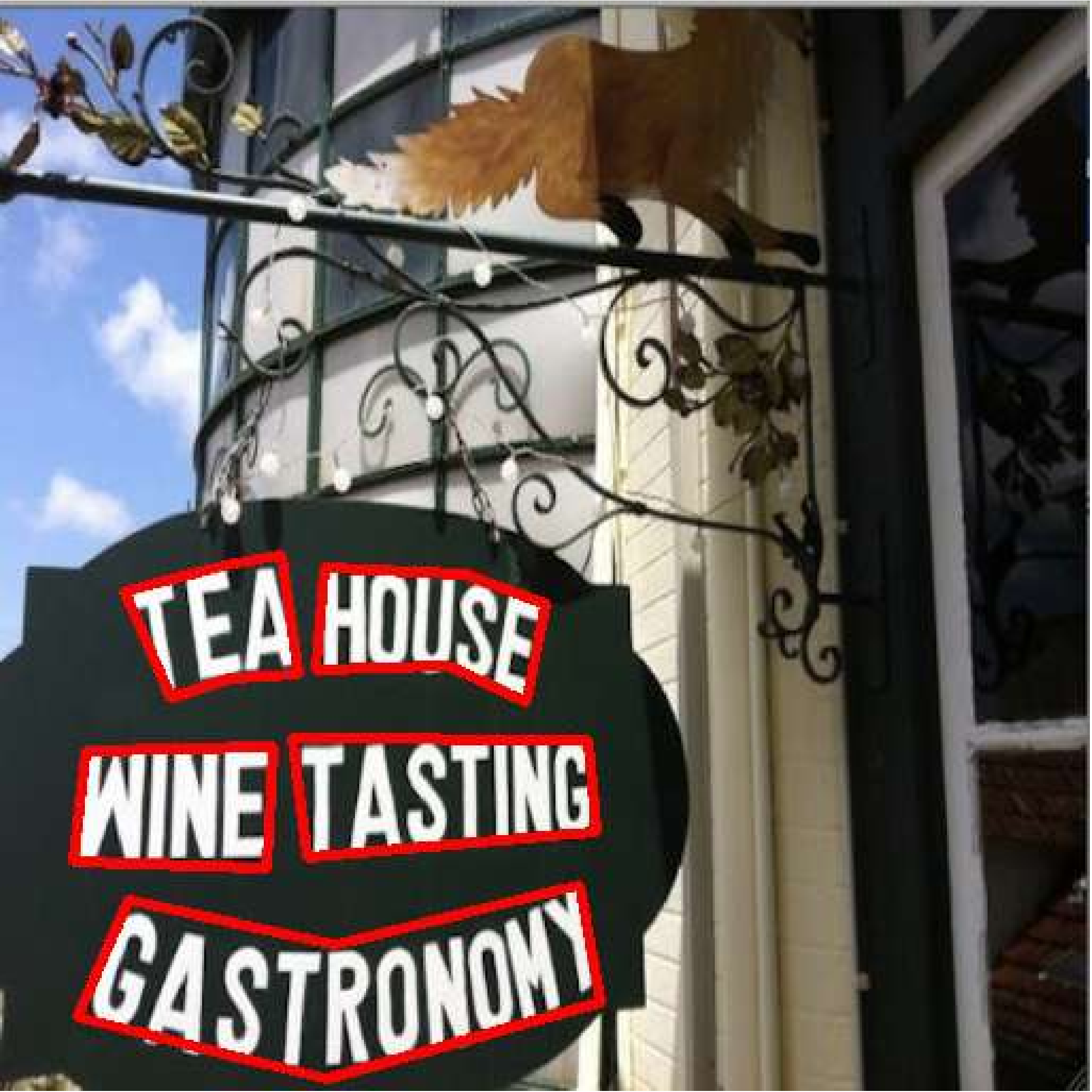}
		\includegraphics[width=0.2\textwidth,height=0.15\textwidth]{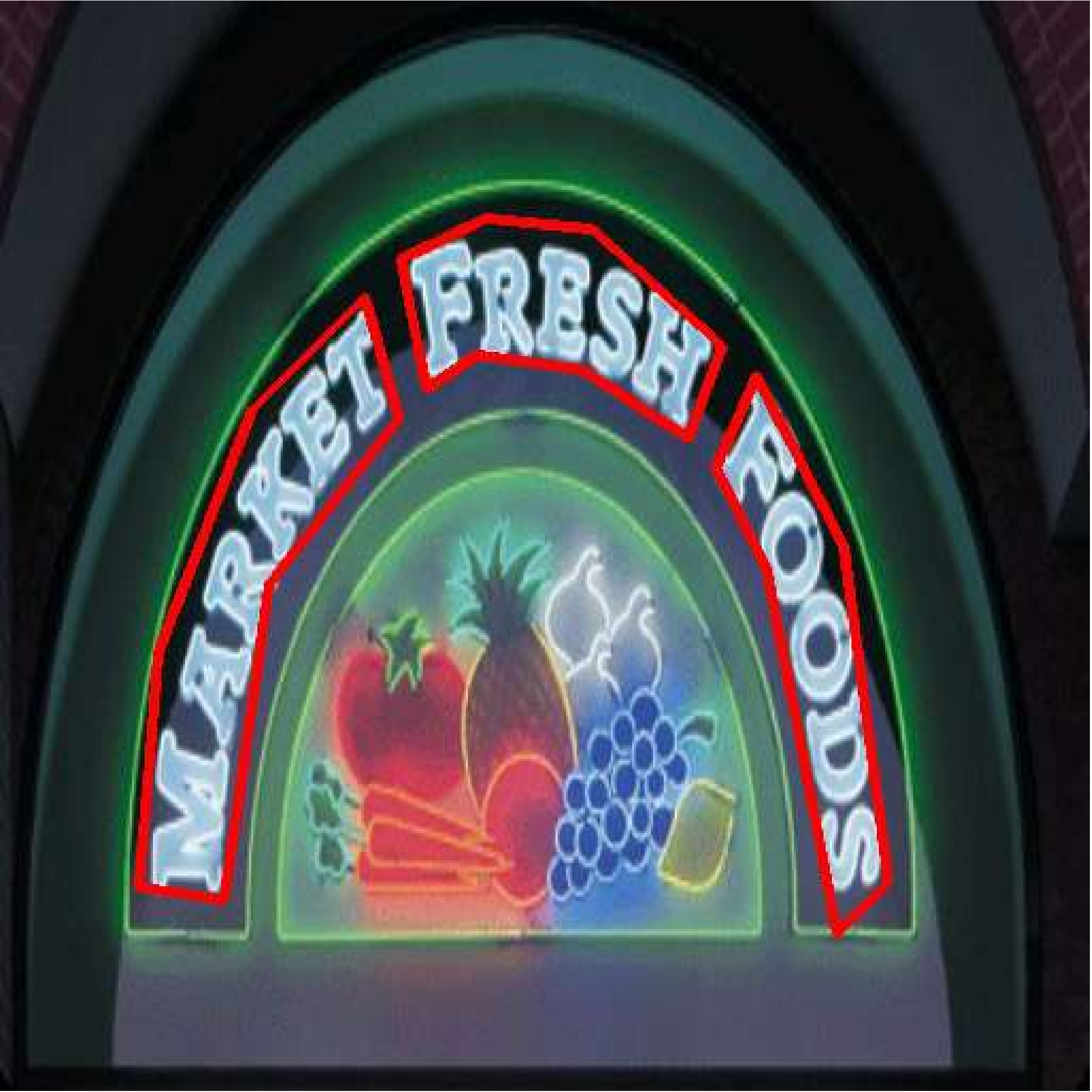}
		\includegraphics[width=0.2\textwidth,height=0.15\textwidth]{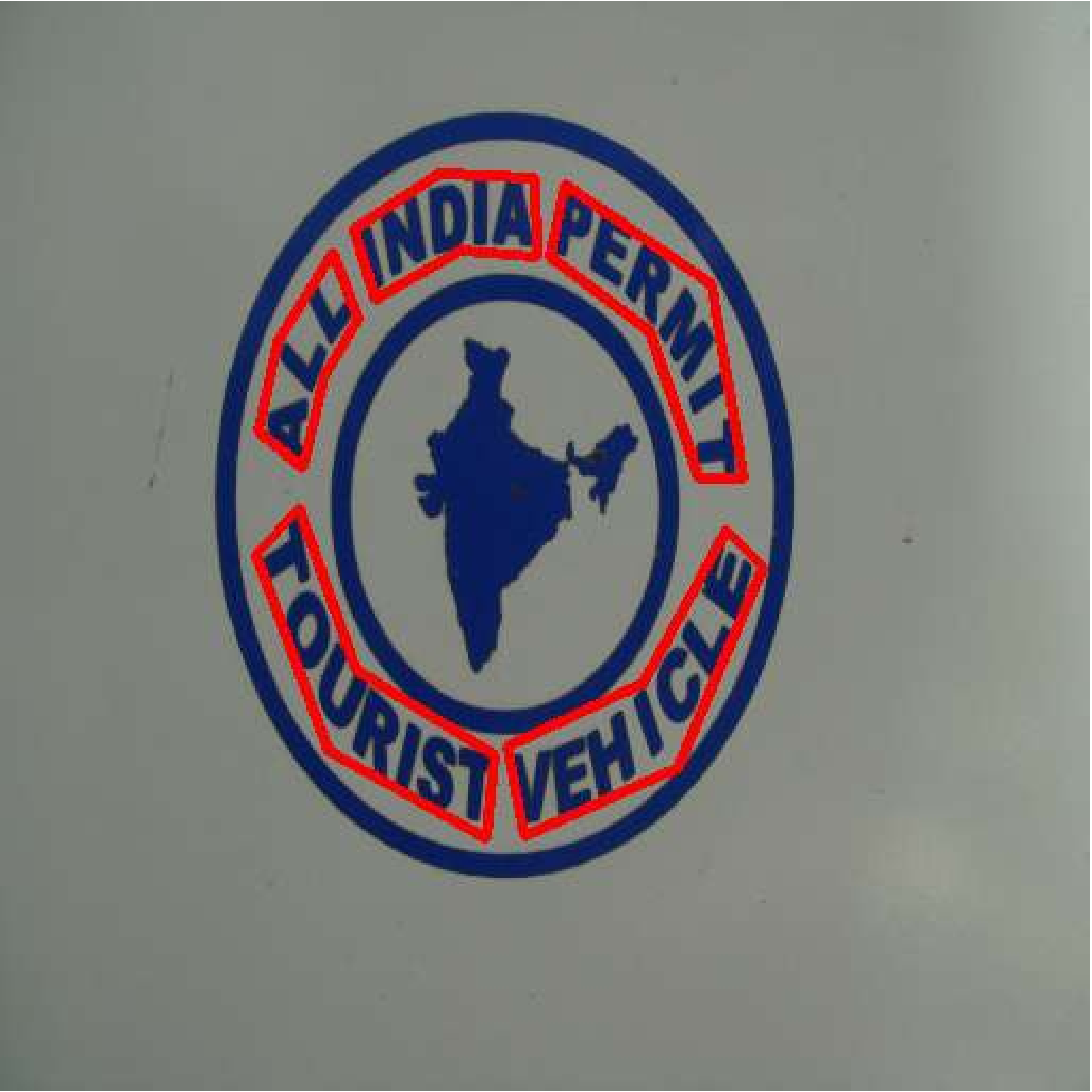}
		\includegraphics[width=0.2\textwidth,height=0.15\textwidth]{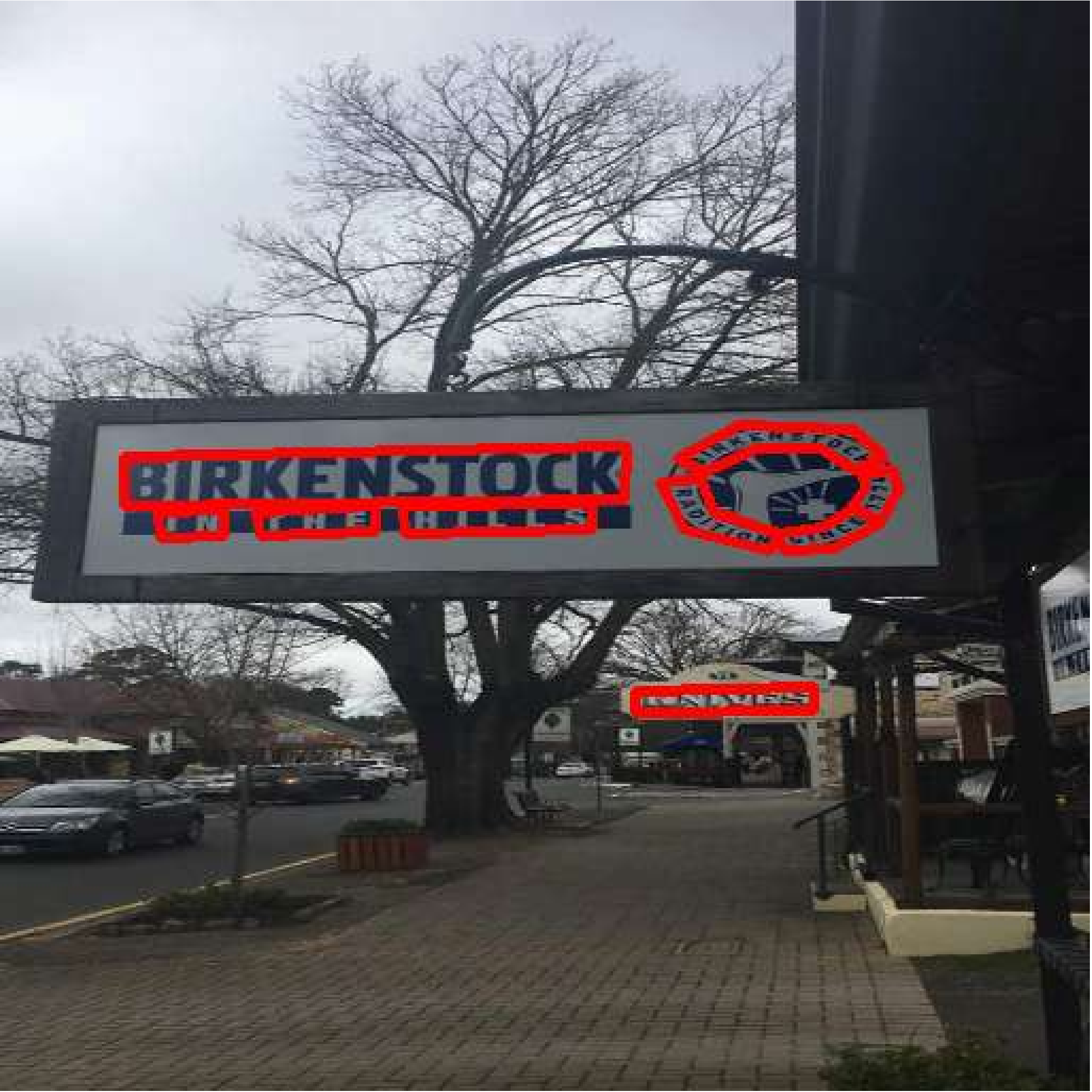}
		\includegraphics[width=0.2\textwidth,height=0.15\textwidth]{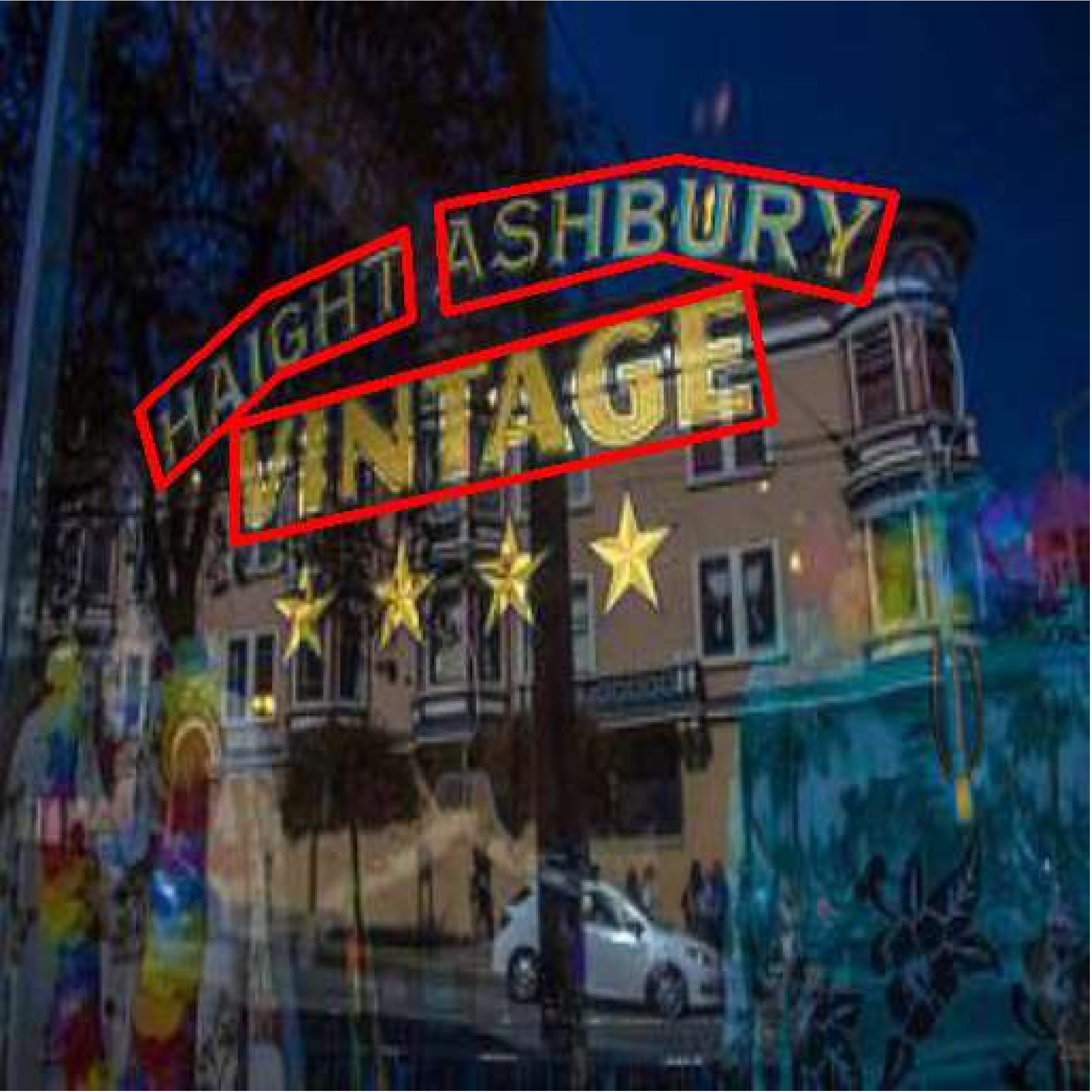}
	}
	
	\subfigure[SCUT-CTW 1500]{
		\includegraphics[width=0.2\textwidth,height=0.15\textwidth]{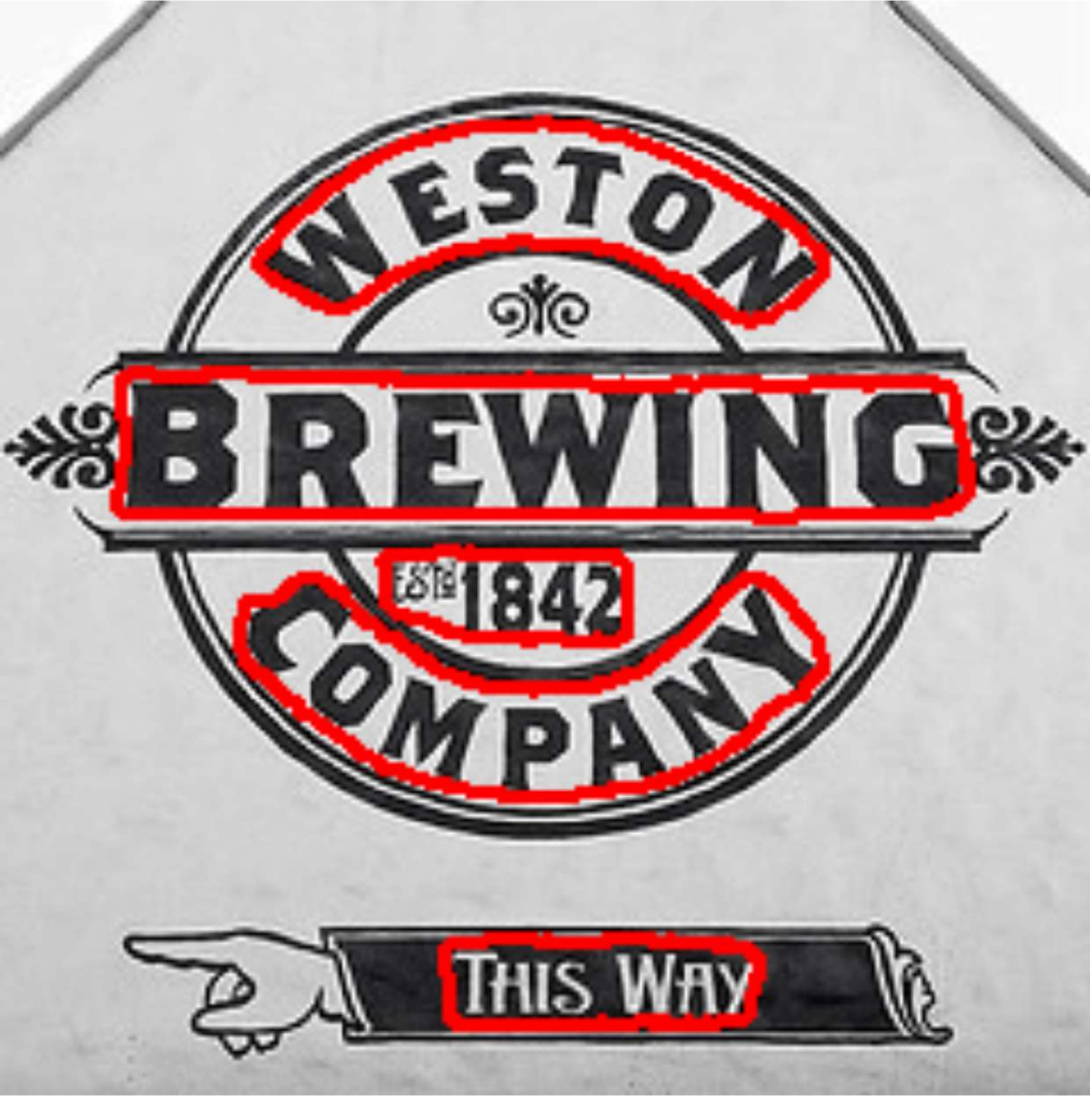}
		\includegraphics[width=0.2\textwidth,height=0.15\textwidth]{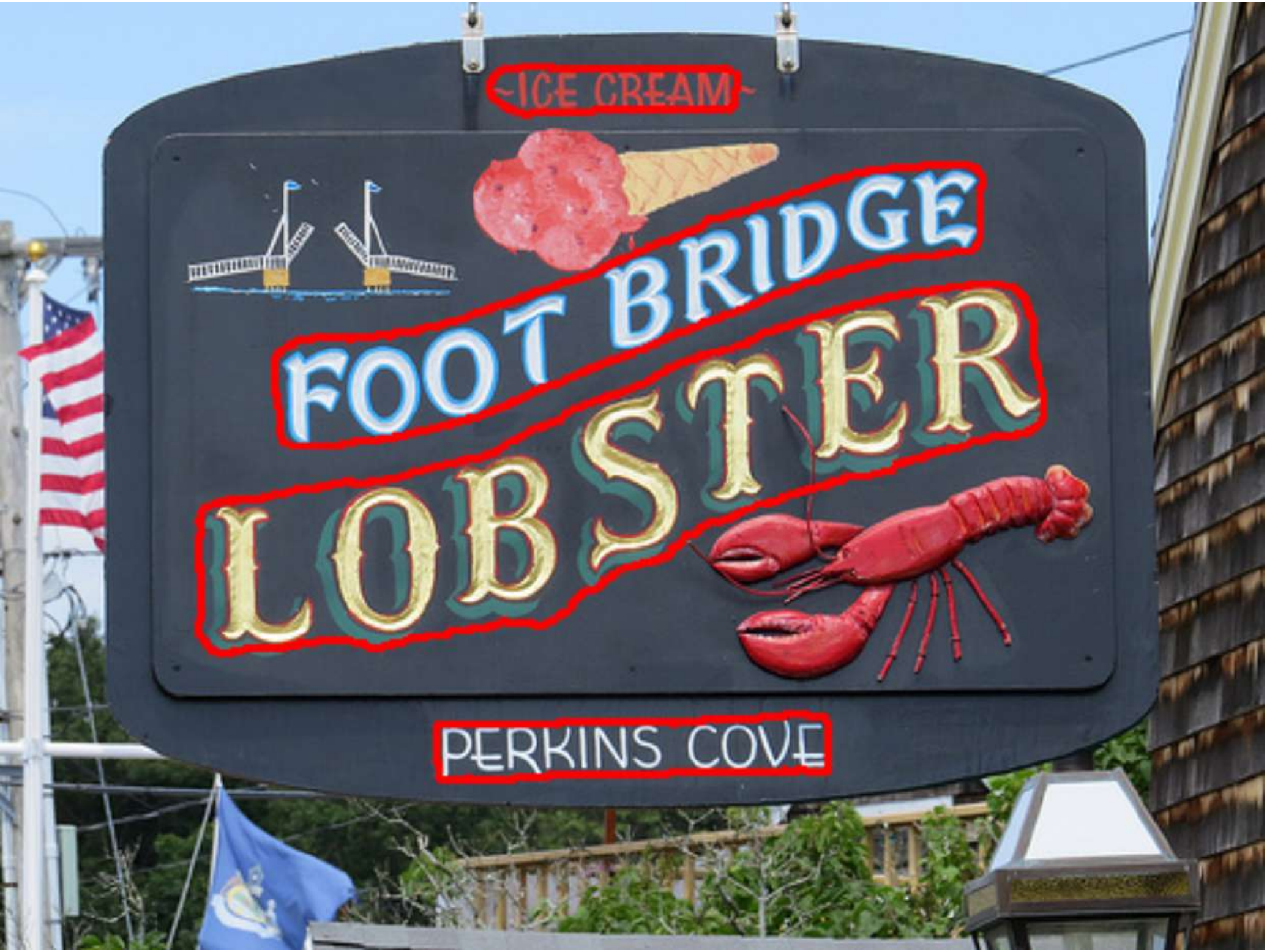}
		\includegraphics[width=0.2\textwidth,height=0.15\textwidth]{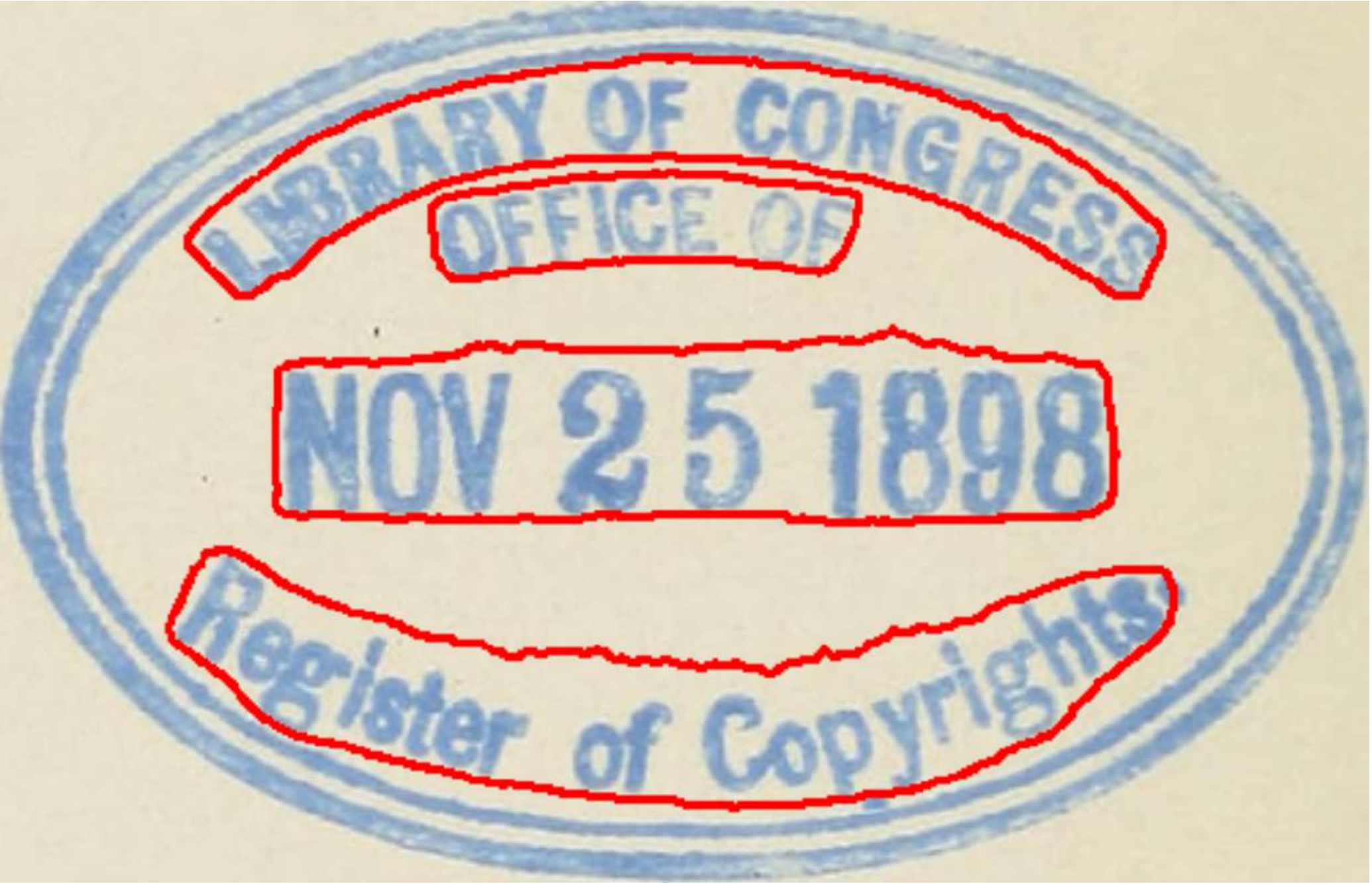}
		\includegraphics[width=0.2\textwidth,height=0.15\textwidth]{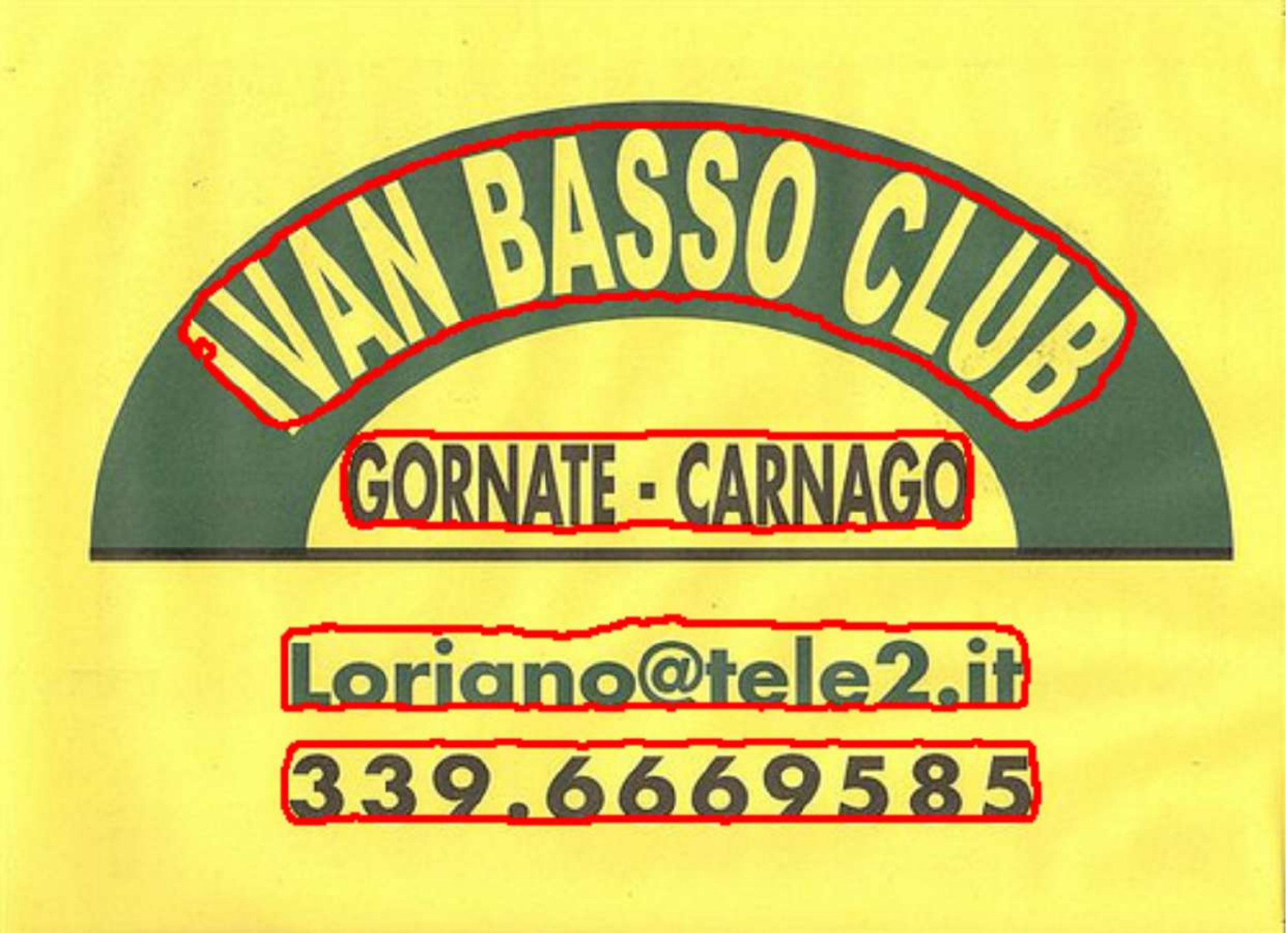}
		\includegraphics[width=0.2\textwidth,height=0.15\textwidth]{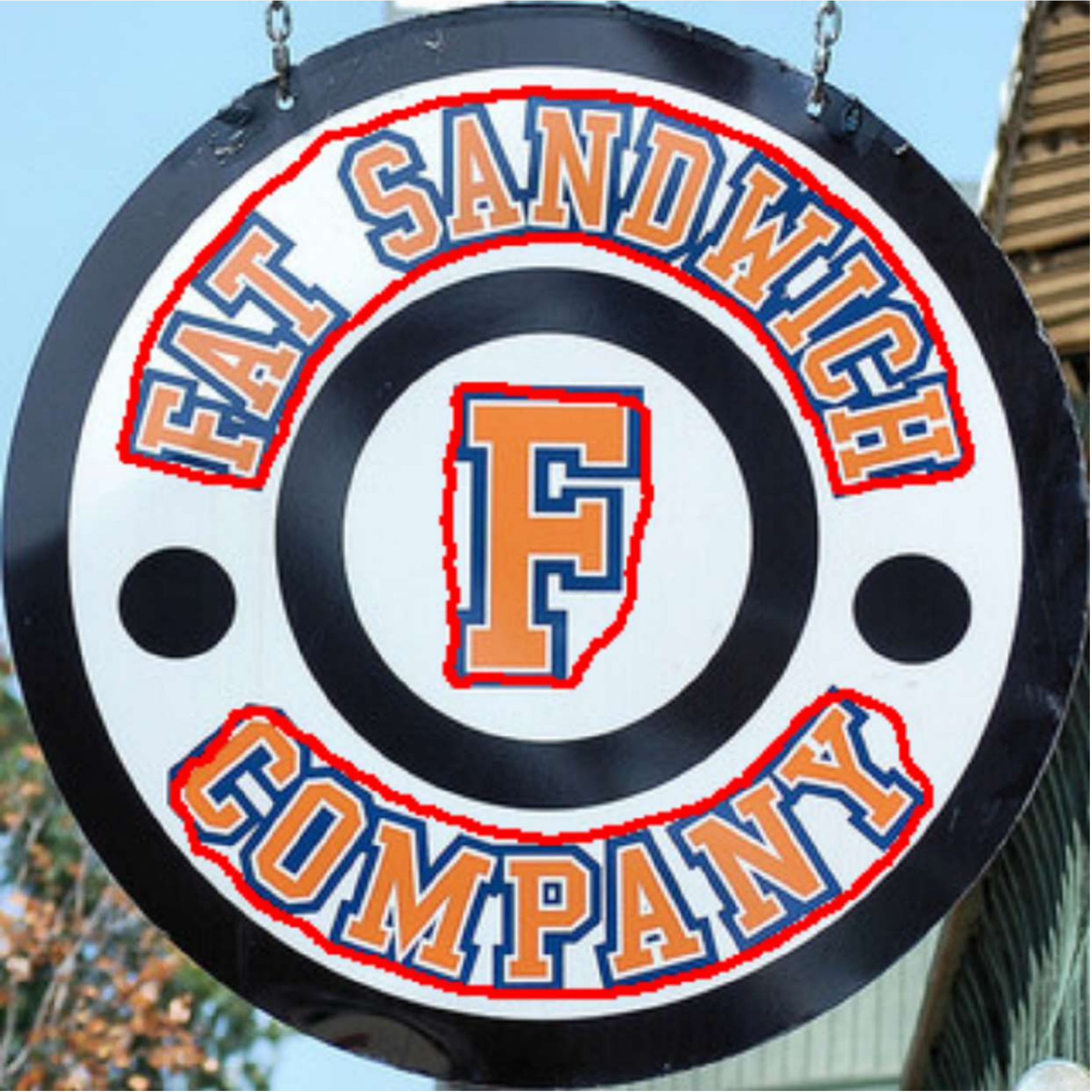}
	}
	
	\subfigure[ICDAR 2015]{
		\includegraphics[width=0.2\textwidth,height=0.15\textwidth]{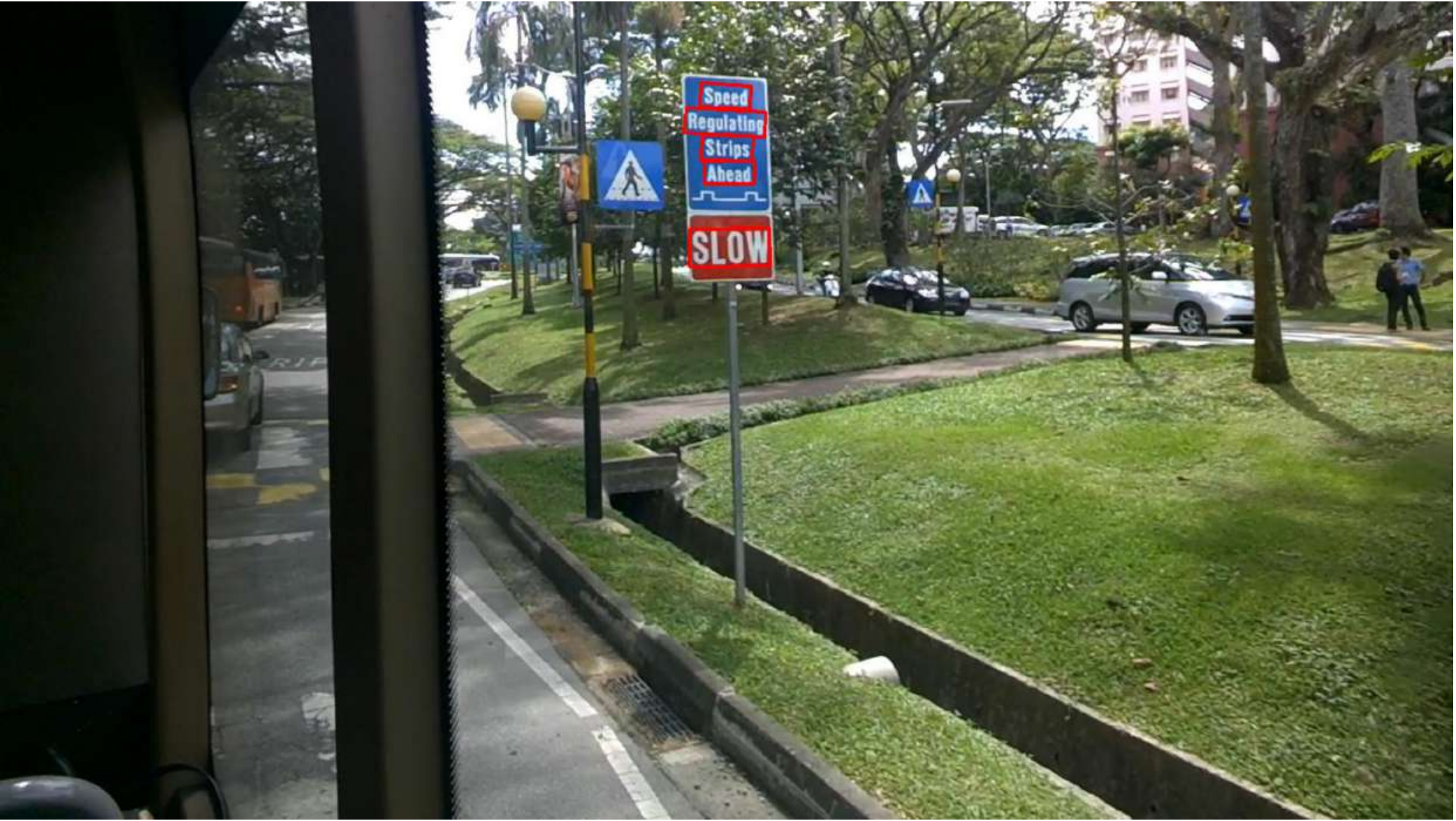}
		\includegraphics[width=0.2\textwidth,height=0.15\textwidth]{experiments/results/img_68.pdf}
		\includegraphics[width=0.2\textwidth,height=0.15\textwidth]{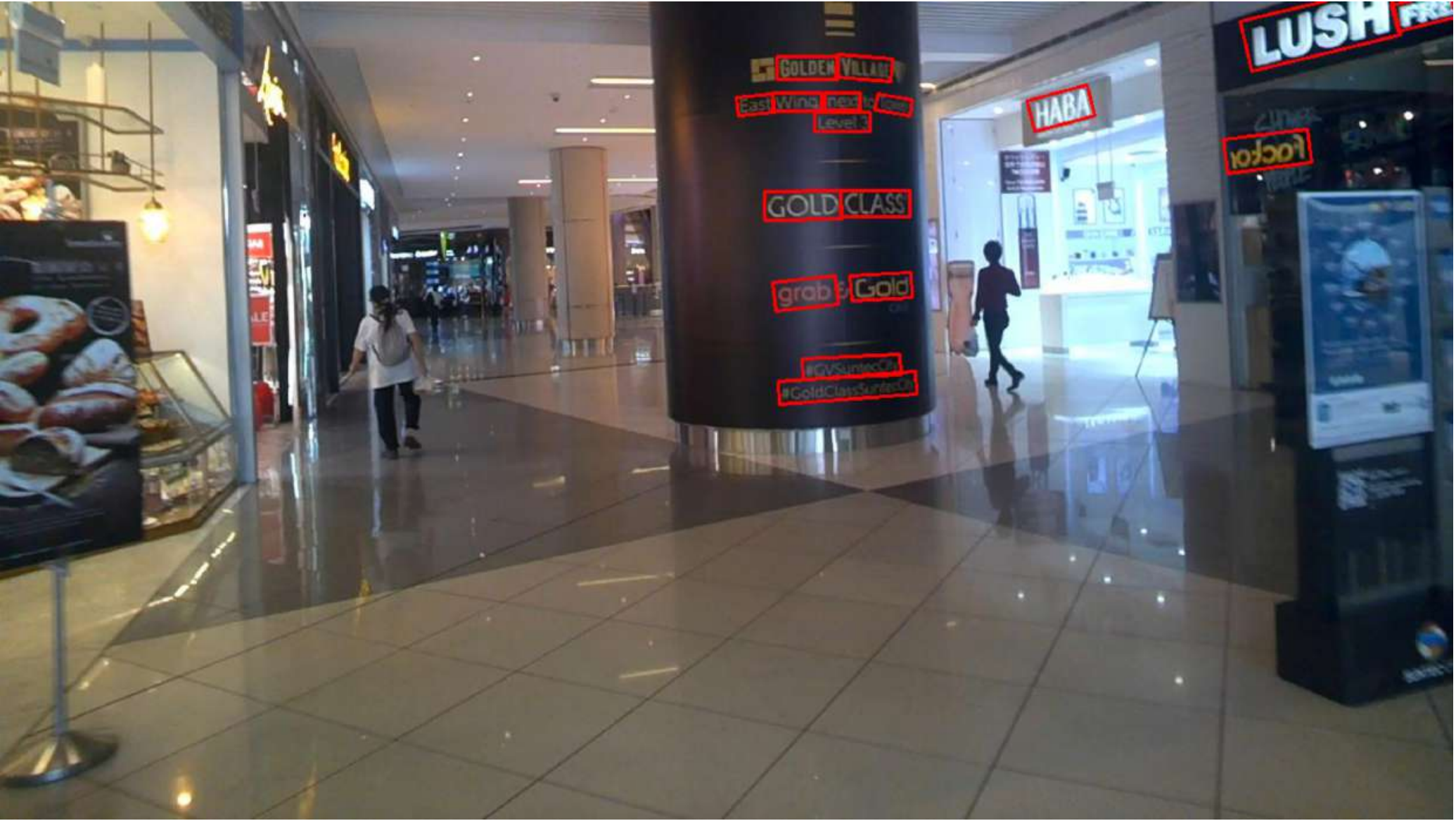}
		\includegraphics[width=0.2\textwidth,height=0.15\textwidth]{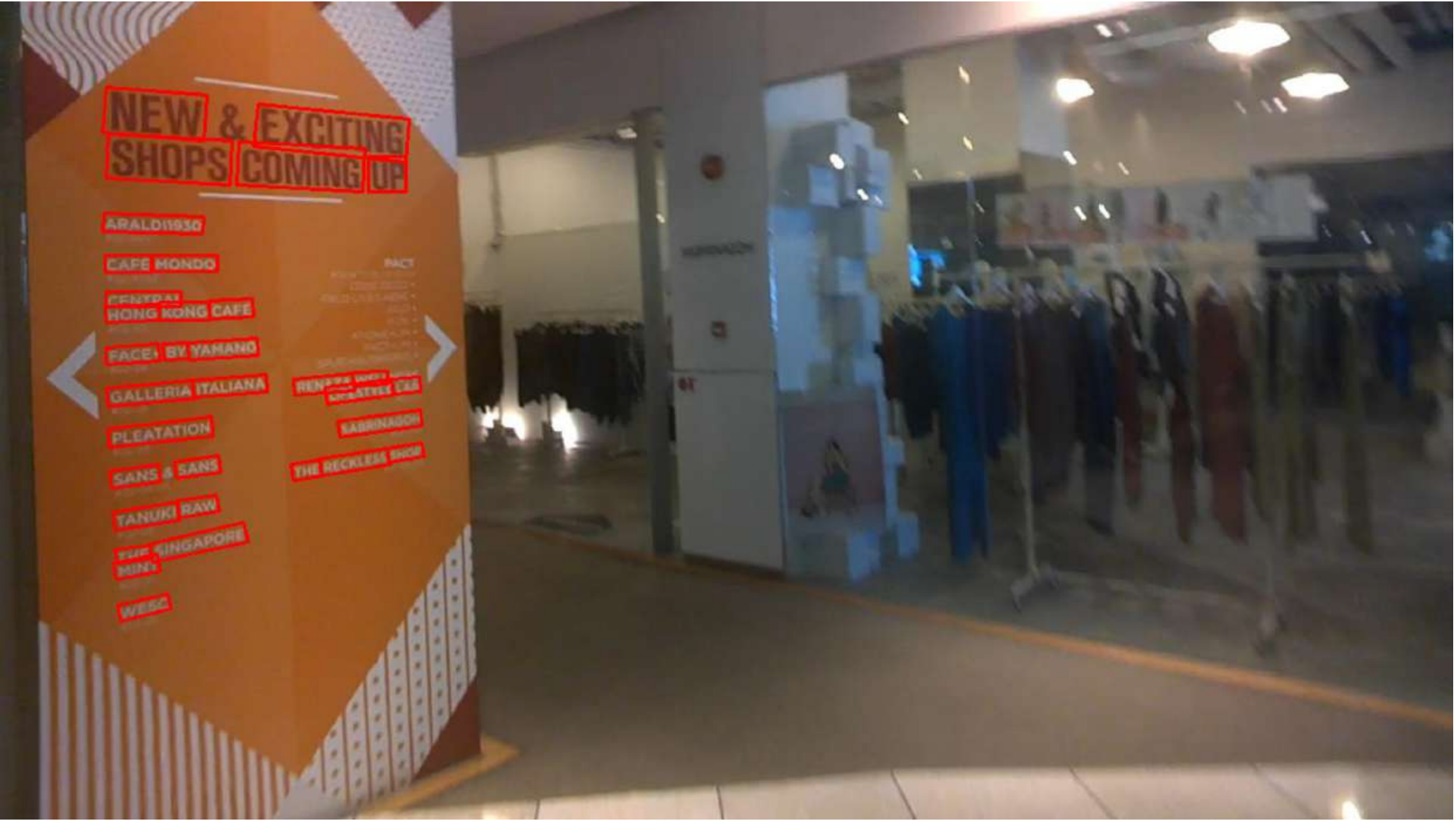}
		\includegraphics[width=0.2\textwidth,height=0.15\textwidth]{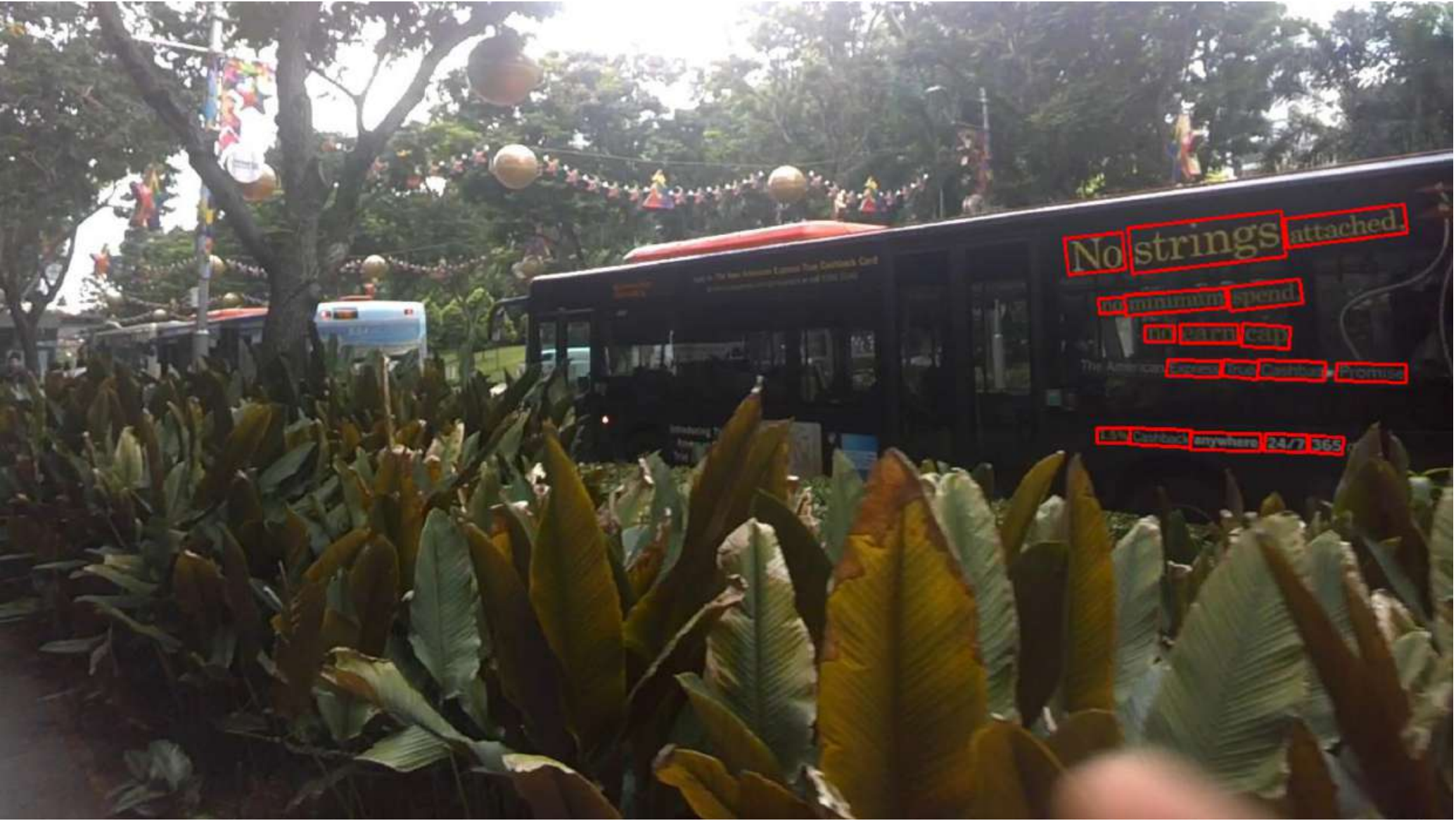}
	}
	\caption{Qualified detection results of Total Text, SCUT-CTW 1500 and ICDAR 2015.}
	\label{fig:exp_res}
	%\vspace{-4pt}
\end{figure*}

\begin{table*}[!ht]
	\caption{Results on Total-Text, SCUT-CTW 1500 and ICDAR 2015 datasets.}
	\label{table:qua_res}
	\centering
	\begin{threeparttable}
		%\setlength{\tabcolsep}{3pt}    
		%\begin{tabular}{|c|c|c|c|c|c|c|c|c|c|c|c|c|}
		\begin{tabular}{ccccccccccccc}
			%\hline
			%backbone & with/without GIM          & Recall   & Precession & f-measure\\
			%\hline
			\toprule[1pt]
			\multirow{2}*{Model} & \multicolumn{4}{c}{Total-Text} & \multicolumn{4}{c}{CTW 1500 } & \multicolumn{4}{c}{ICDAR 2015}\\
			\cline{2-13}
			~                    & \textit{R} & \textit{P} & \textit{H}  & \textit{F} & \textit{R} & \textit{P} & \textit{H}  & \textit{F} & \textit{R} & \textit{P} & \textit{H}  & \textit{F}\\
			%\hline
			\toprule[1pt]
			%EAST\cite{zhou2017east} & 36.2 & 50.0  & 42.0 & - & - & - & - & - \\
			Deconv~\cite{ch2017total} & 40.0 & 33.0  & 36.0 & -  & - & - & - & - & - & - & - & -\\
			TextField~\cite{xu2019textfield} & 79.9 & 81.2 & 80.6 & - & -  & - & - & - & 80.5 & 84.3 & 82.4 & \textbf{6.0}\\
			Wang \textit{et al.}~\cite{wang2019all} & 83.5 & 85.2 & 84.3 & - & - & - & - & - & 82.2 & 88.1 & 85.0 & -\\
			CTPN~\cite{tian2016detecting} & -  & - & - & - & 53.8 & 60.4 & 56.9 & 7.14 & 52.0 & 74.0 & 61.0 & -\\
			CTD~\cite{yuliang2017detecting} & -  & - & - & - & 69.8 & 77.4 & 73.4 & 13.3 & - & - & - & -\\
			SLPR~\cite{zhu2018sliding} & -  & - & - & - &  70.1 & 80.1 & 74.8 & -& 83.6 & 85.5 & 84.5  & -\\
			SegLink~\cite{shi2017detecting} & 23.8 & 30.3  & 26.7 & - & 40.0 & 42.3 & 40.8 & 10.7 & 76.8 & 73.1 & 75.0 & - \\
			EAST~\cite{zhou2017east} & 36.2 & 50.0  & 42.0 & -  & 49.1 & 78.7 & 60.4 & \textbf{21.2} & 78.3 & 83.3 & 80.7 & -\\
			PSENet~\cite{li2018shape} & 75.1 & 81.8  & 78.3 & 3.9 & 75.6 & 80.6 & 78.0 & 3.9 & 79.7 & 81.5 & 80.6 & 1.6\\
			TextSnake~\cite{long2018textsnake} & 74.5 & 82.7 & 78.4 & - & \textbf{85.3} & 67.9 & 75.6  &-& 80.4 & 84.9 & 82.6 & 1.1\\
			LOMO~\cite{zhang2019look} & 69.6 & \textbf{89.2} & 78.4 & - & 75.7 & \textbf{88.6} & 81.6 & - & 83.5 & \textbf{91.3} & 87.2 & -\\
			ABCNet~\cite{liu2020abcnet} & - & - & 78.4 & - & - & - & 74.1 & - & - & - & - & -\\
			TextRay~\cite{wang2020textray} & 77.9 & 83.5 & 80.6 & - & 80.4 & 82.8 & 81.6 & - & - & - & - & -\\
			$\text{NASK}_{conf}$~\cite{cao2020all} & 81.2 & 83.3 &  82.2  & 8.4  & 78.3 & 82.8 &  80.5 & 12.1 & 86.8 & 90.2 & 88.5 & 4.2\\ 
			%NASK+SE & 82.7 & 84.2 & 83.4  & 8.4  & 78.3 & 82.8 & 80.5 & 12.1 & - & - & - & -\\ 
			NASK & \textbf{83.2} &  85.6 &  \textbf{84.4}  & \textbf{8.4}  & 80.1 & 83.4 & \textbf{81.7} & 12.1 & \textbf{89.2} & 90.9 & \textbf{90.0} & 4.2\\ 
			\hline
			\toprule[1pt]
		\end{tabular} 
		\begin{tablenotes}
			\footnotesize
			\item Note: \textit{R}, \textit{P}, \textit{H}, \textit{F} denote Recall, Precision, Hmean and FPS respectively. $\text{NASK}_{conf}$ is our previous conference version~\cite{cao2020all}. All data are given in percentile form.
			%\item[2] .
		\end{tablenotes}
	\end{threeparttable}
\end{table*}

\begin{table*}[!ht]
	\caption{Ablation studies on SCUT-CTW 1500.}
	\label{table:ablation}
	\centering
	\begin{threeparttable}
		
		%\begin{tabular}{|c|c|c|c|c|c|c|c|c|c|}
		\begin{tabular}{cccccccccc}
			\hline
			\toprule[1pt]
			Experiment & \(1^{st} seg\) & \(2^{nd} seg\) & \textit{Attention} & \textit{Pooling} & \textit{\(G\)} & \textit{R} & \textit{P} & \textit{H} & \textit{F} \\
			\hline
			\toprule[1pt]
			\multirow{5}*{(a)} & \multirow{5}*{\ding{51}} &\multirow{5}*{\ding{51}} &\multirow{5}*{GSCA} & \multirow{5}*{GeoAlign} & 2 & 80.8 & 83.8 & 82.3 & 3.4 \\
			%~ & ~ & ~ & ~ & ~  & 2 & 80.8 & 83.8 & 81.2 & 3.4 \\
			%~ & ~ & ~ & ~ & ~  & 4 & 78.3 & 82.8 & 80.5 & 12.1 \\
			~ & ~ & ~ & ~ & ~ & 4 & 80.1 & 83.4 & 81.7 & 12.1 \\
			~ & ~ & ~ & ~ & ~ & 8 & 79.2 & 82.2 & 80.7 & 12.9 \\
			~ & ~ & ~ & ~ & ~ & 12 & 78.7 & 81.8 & 80.2 & 13.7 \\
			~ & ~ & ~ & ~ & ~ & 16 & 78.2 & 81.0 & 79.6 & 13.8 \\
			\hline
			\multirow{3}*{(b)} & \ding{55} & \ding{51} & \multirow{3}*{GSCA} & \multirow{3}*{GeoAlign} & \multirow{3}*{4} & 75.2 & 76.4 & 75.8 & 14.7 \\
			~ & \ding{51} & \ding{55} & ~ & ~ & ~ & 73.1 & 72.4 & 72.7 & 18.8 \\
			~ & \ding{51} & \ding{51} & ~  & ~ & ~ & 80.1 & 83.4 & 81.7 & 12.1 \\
			\hline
			\multirow{5}*{(c)} & \multirow{5}*{\ding{51}} & \multirow{5}*{\ding{51}} & GSCA &  \multirow{5}*{GeoAlign} & \multirow{2}*{4} & 80.1 & 83.4 & 81.7 & 12.1 \\
			~ & ~ & ~ & $\text{GSCA}_{conf}$  & ~ & ~ & 78.9 & 82.5 & 80.7 & 12.1 \\
			~ & ~ & ~ & DANet & ~ & - & 79.8 & 83.1 & 81.4 & 10.3 \\
			~ & ~ & ~ & CCNet & ~ & - & 78.7 & 82.4 & 80.5 & 13.8 \\
			~ & ~ & ~ & None & ~ & - & 78.2 & 81.1 & 79.5 & 16.7 \\
			\hline 
			\multirow{3}*{(d)} & \multirow{3}*{\ding{51}} & \multirow{3}*{\ding{51}} & \multirow{3}*{GSCA}  & GeoAlign  &  \multirow{3}*{4} & 80.1 & 83.4 & 81.7 & 13.1 \\
			~ & ~ & ~ & ~ & RoI Align & ~ & 79.2 & 82.4 & 80.8 & 13.6 \\
			~ & ~ & ~ & ~ & RoI Pooling & ~ & 78.5 & 81.7 & 80.1 & 14.3 \\
			%\hline 
			\toprule[1pt]
		\end{tabular} 
		\begin{tablenotes}
			\footnotesize
			\item Note: $1^{st} seg$ and $2^{nd} seg$ means the first and the second stage segmentation network;  \textit{Attention} denotes different attention modules including GSCA, $\text{GSCA}_{conf}$, DANet, and CCNet; $\text{GSCA}_{conf}$ denotes the preceding version of GSCA in the conference paper~\cite{cao2020all}. \textit{Pooling} denotes the adopted pooling methods including our proposed GeoAlign, RoI Align, and RoI Pooling. $G$ denotes the group number of GSCA and $\text{GSCA}_{conf}$. All data are given in percentile form.%\textit{R},\textit{P},\textit{H},\textit{F} is the same as Table 1.
			%\item[2] .
		\end{tablenotes}
	\end{threeparttable}
\end{table*}

\begin{figure*}[htbp]
	\centering
	\subfigure[Visualization of GSCA feature map]{
		\begin{minipage}{\textwidth}
			\centering
			\includegraphics[width=0.18\textwidth,height=0.15\textwidth]{experiments/ablation/cross_img629.pdf} \vspace{4pt}
			\includegraphics[width=0.18\textwidth,height=0.15\textwidth]{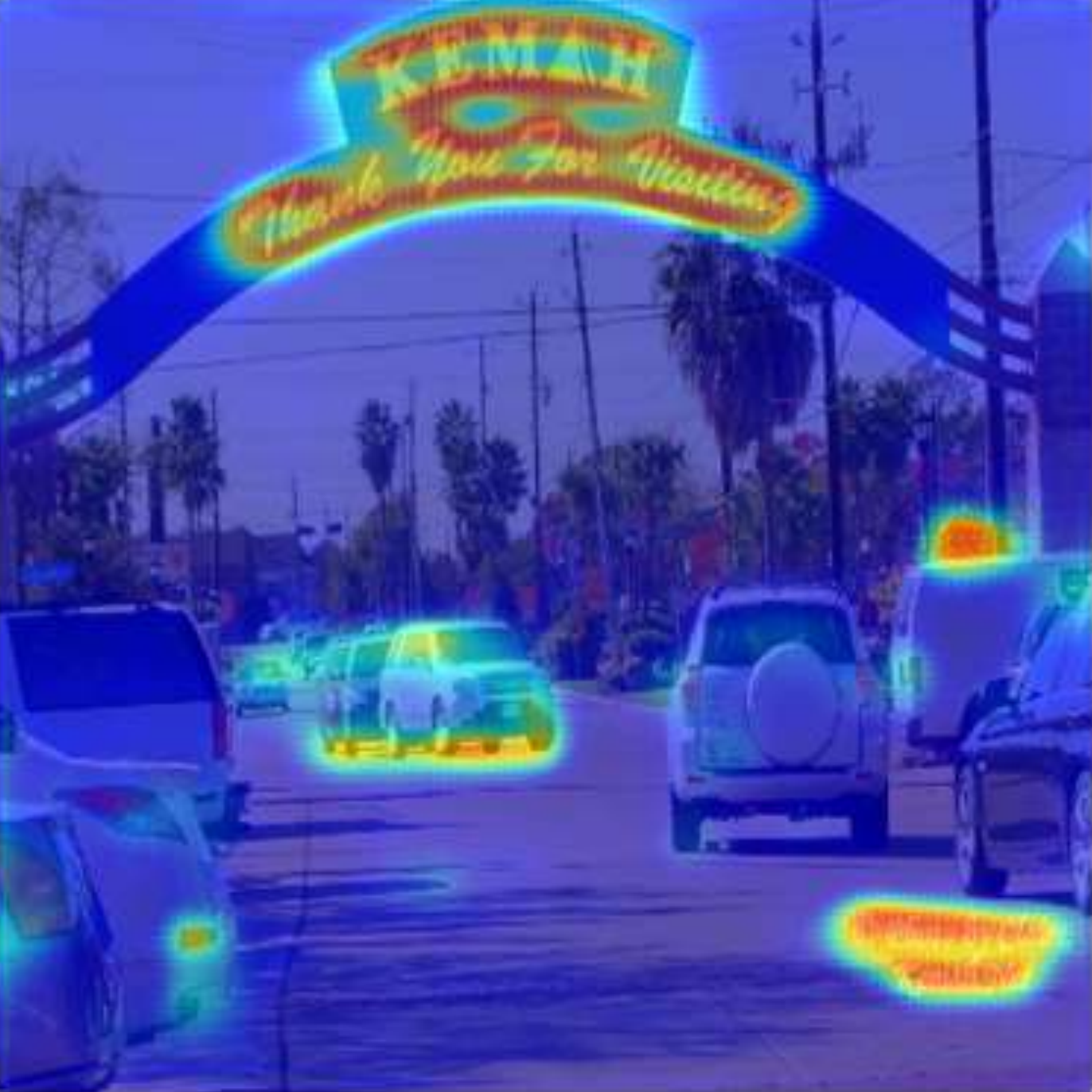} 
			\includegraphics[width=0.18\textwidth,height=0.15\textwidth]{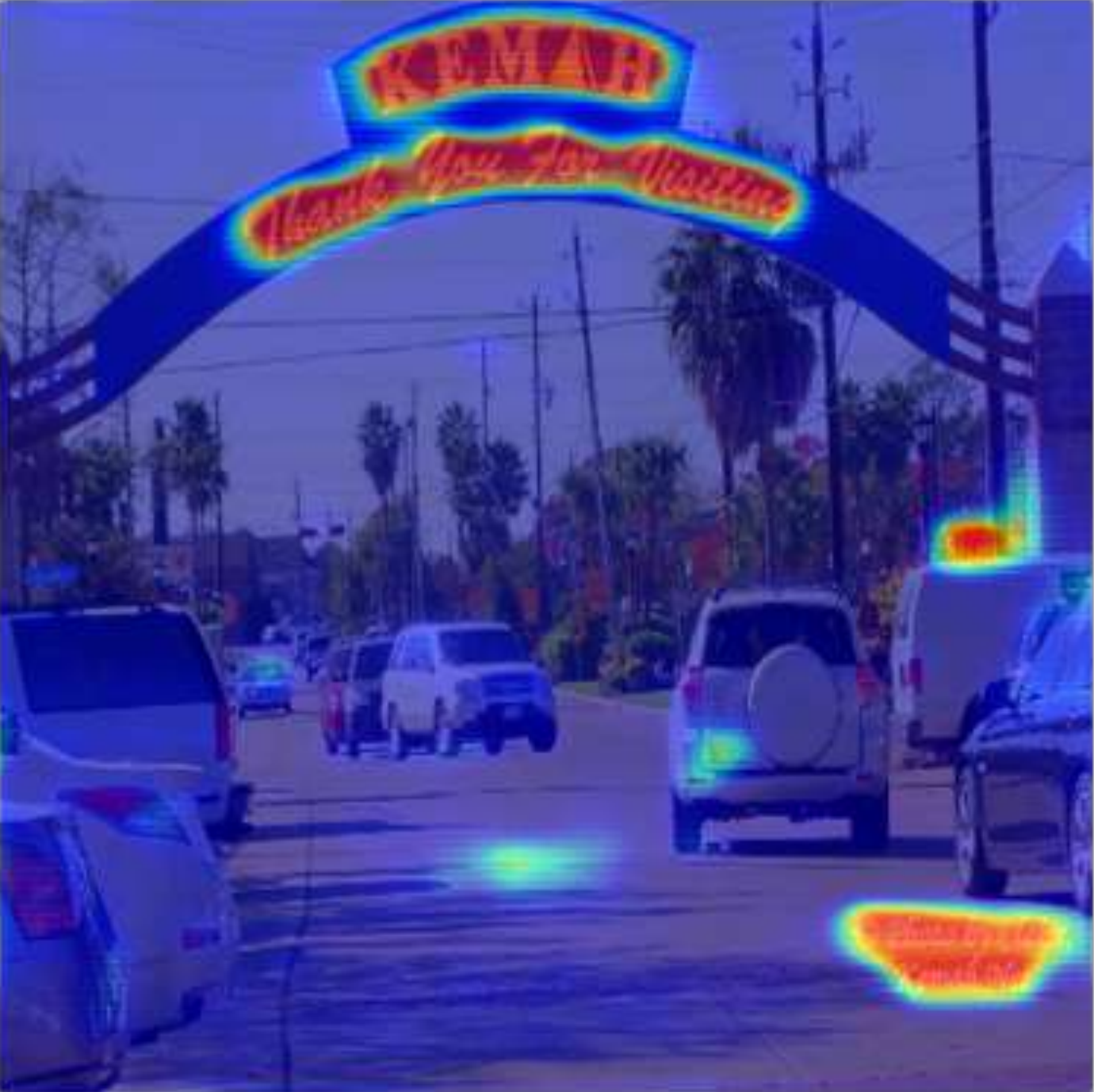}
			\includegraphics[width=0.18\textwidth,height=0.15\textwidth]{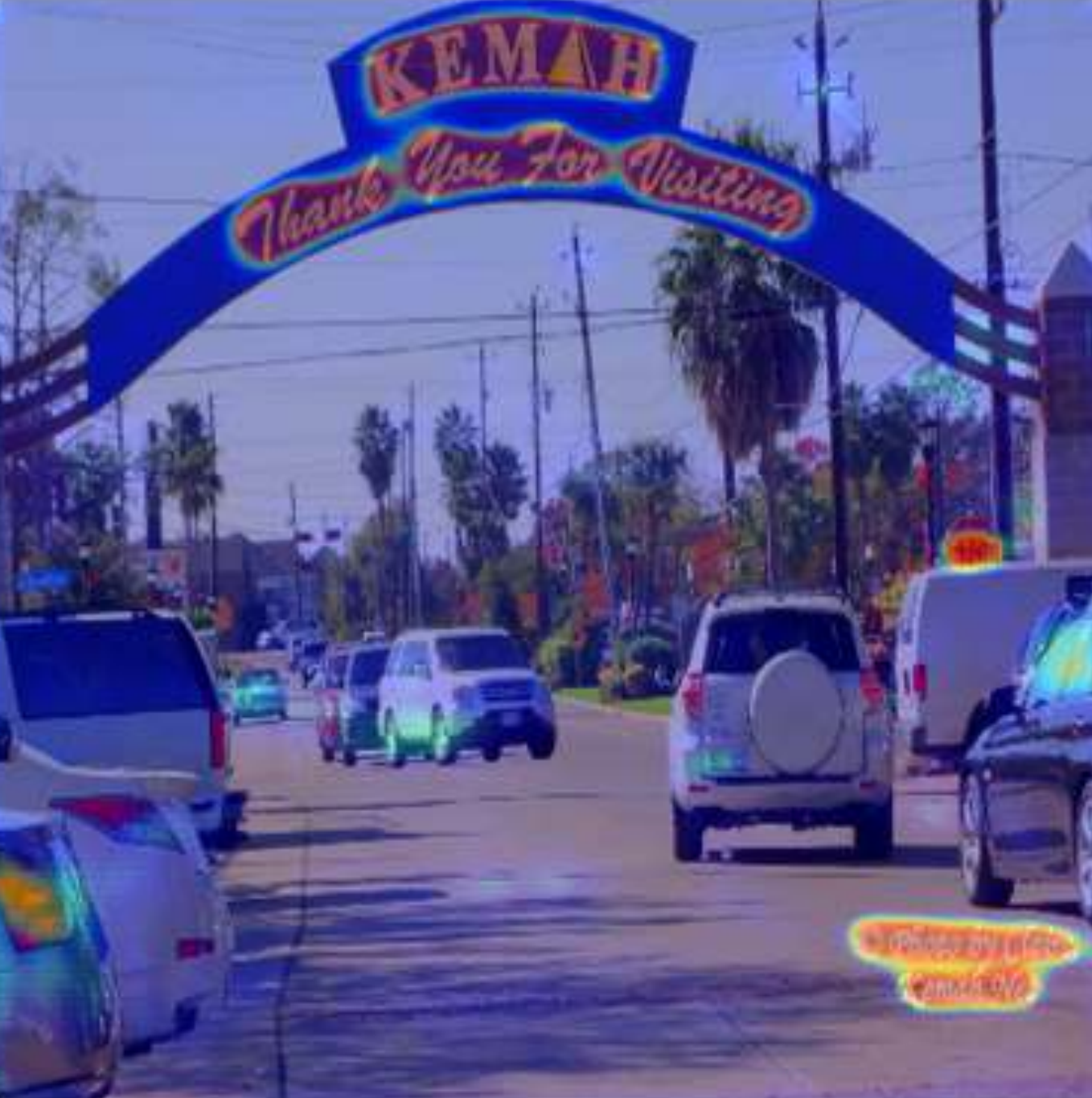}
			\includegraphics[width=0.18\textwidth,height=0.15\textwidth]{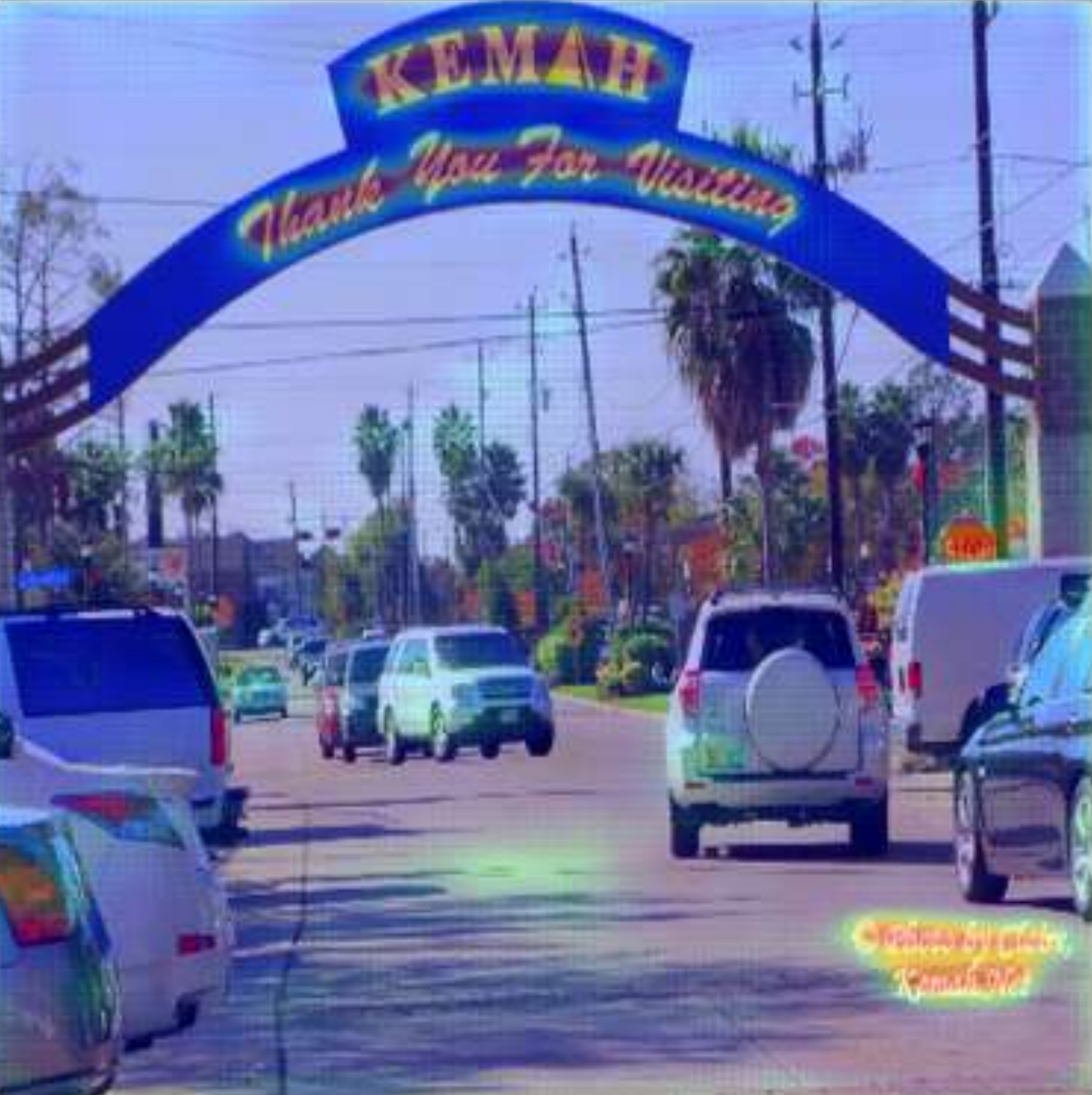} \\
			\includegraphics[width=0.18\textwidth,height=0.15\textwidth]{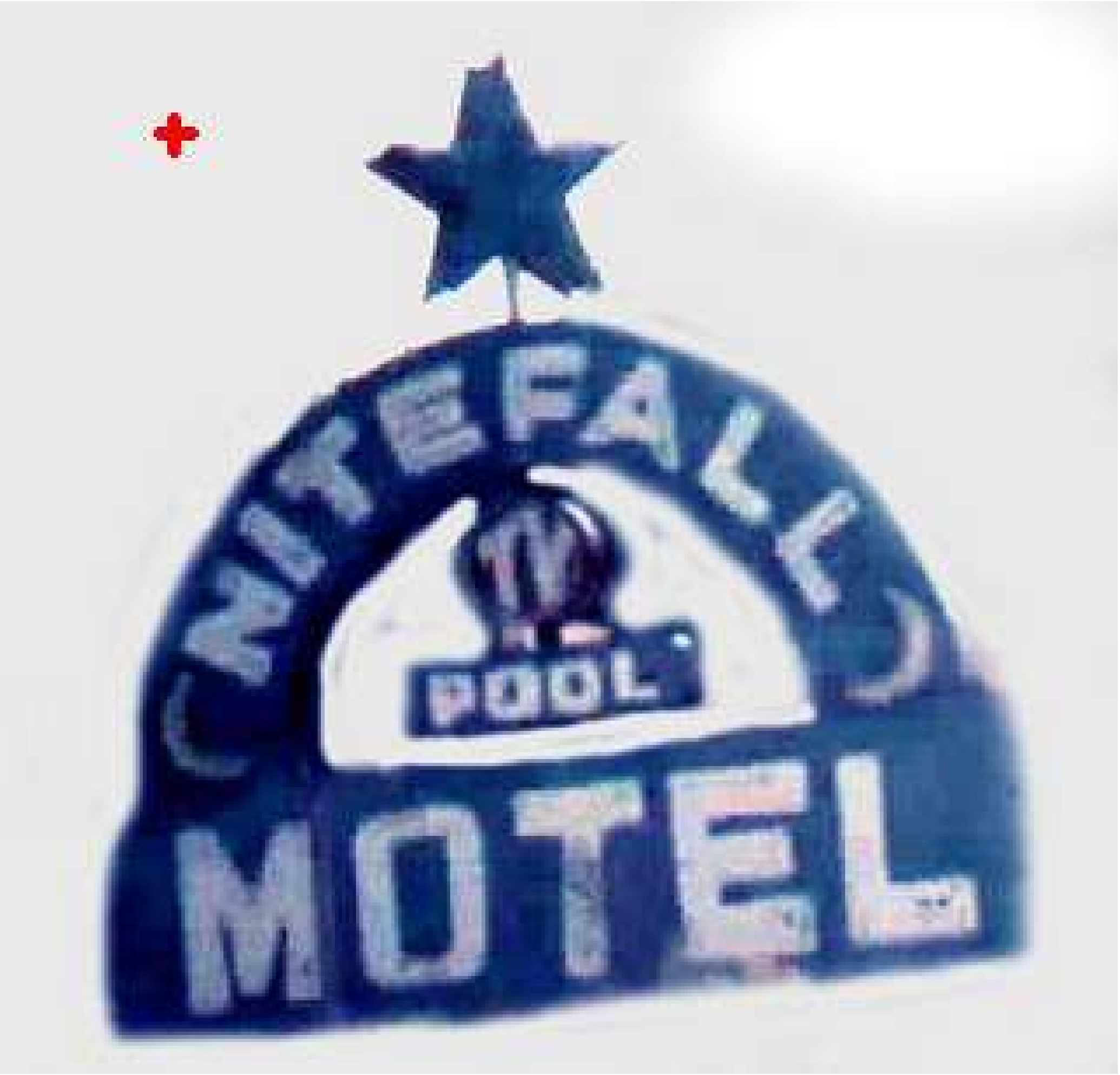} \vspace{5pt}
			\includegraphics[width=0.18\textwidth,height=0.15\textwidth]{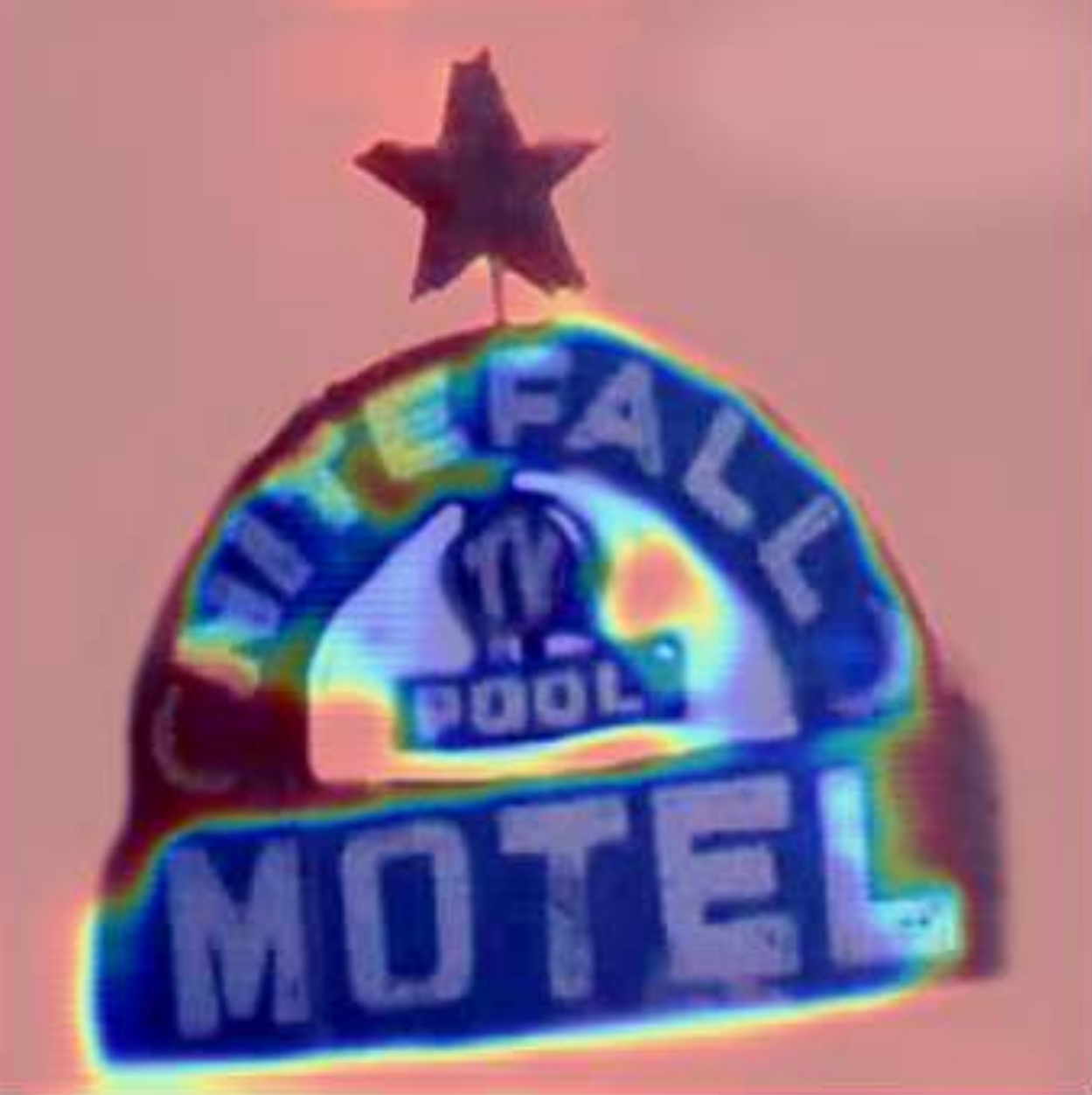}
			\includegraphics[width=0.18\textwidth,height=0.15\textwidth]{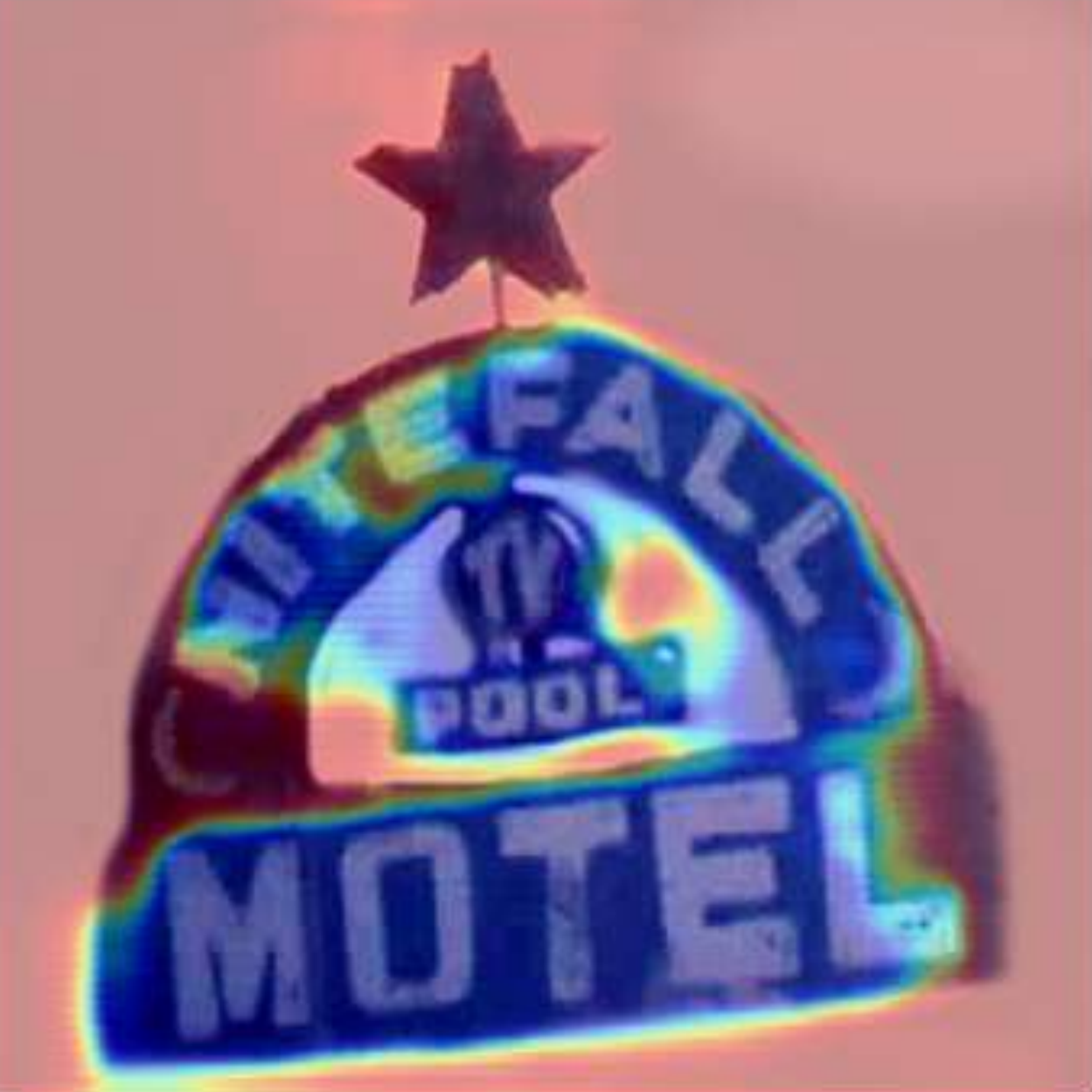}
			\includegraphics[width=0.18\textwidth,height=0.15\textwidth]{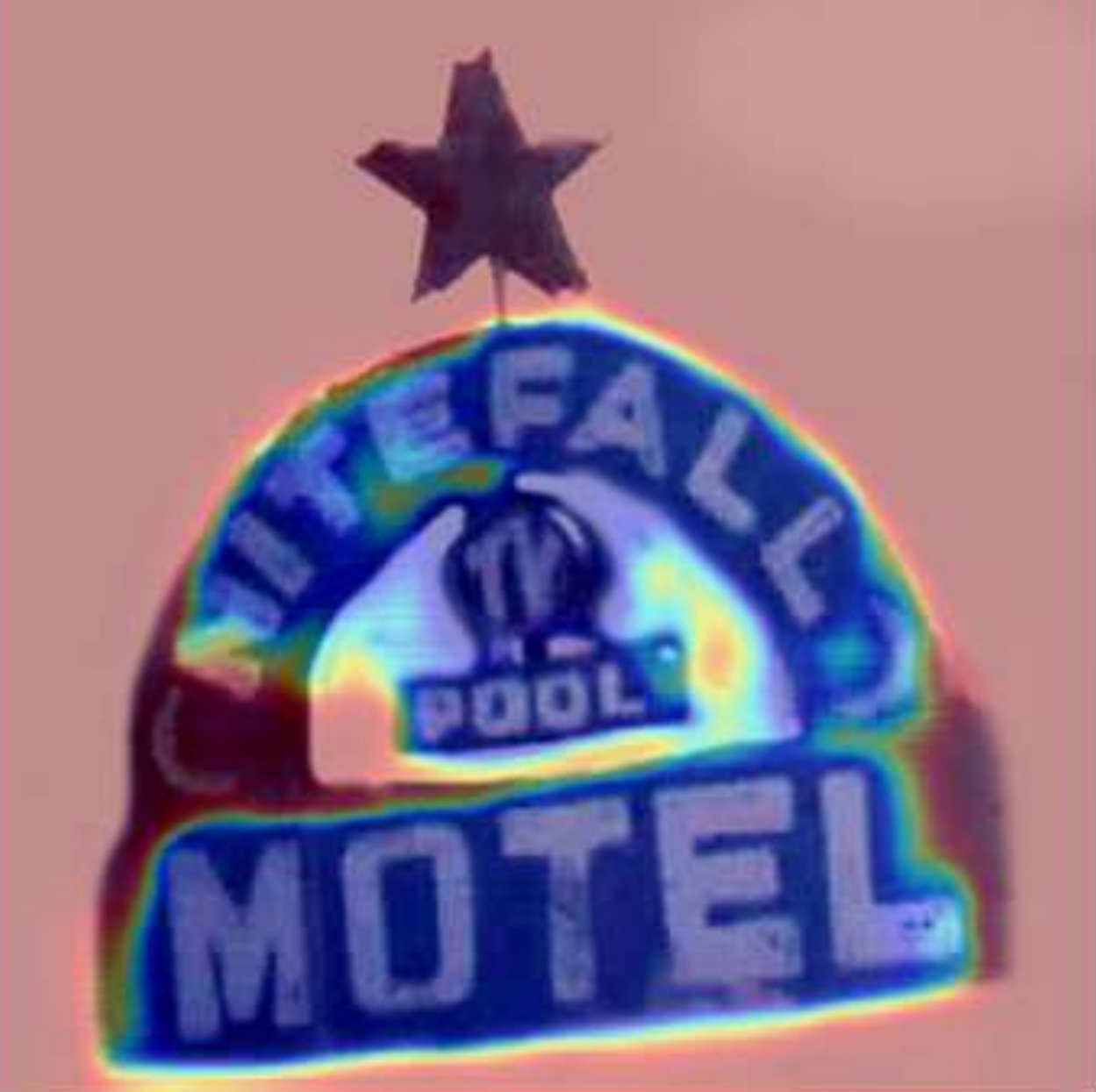}
			\includegraphics[width=0.18\textwidth,height=0.15\textwidth]{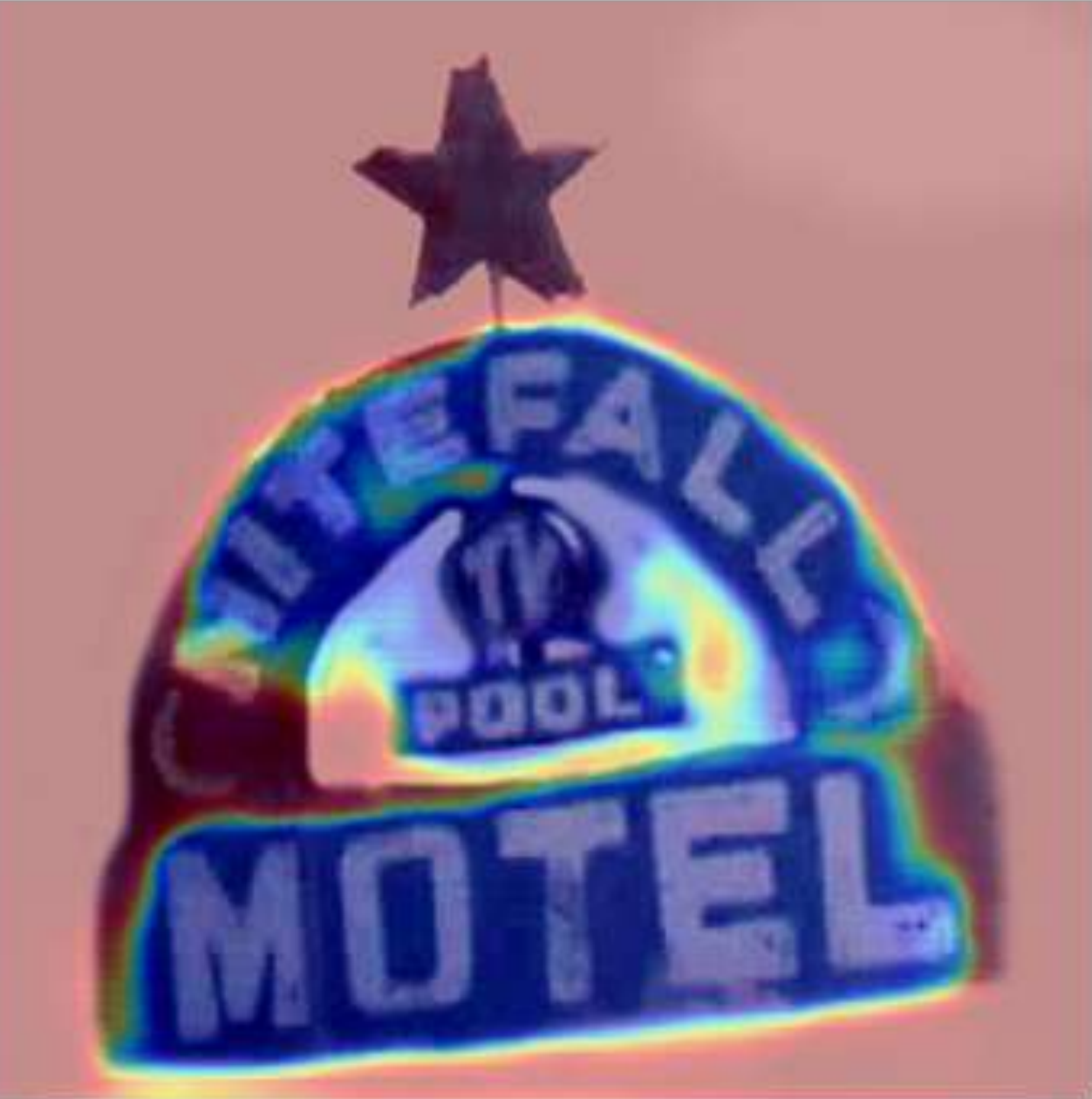}
			%\caption{World Map}
		\end{minipage}
	}
	
	%\subfigure[Visualization of GSCA feature map]{
	%   \includegraphics[width=0.2\textwidth,height=0.15\textwidth]{1296.pdf}
	%   \includegraphics[width=0.2\textwidth,height=0.15\textwidth]{1296tr_grey_epoch80.pdf}
	%   \includegraphics[width=0.2\textwidth,height=0.15\textwidth]{1296tr_grey_epoch100.pdf}
	%   \includegraphics[width=0.2\textwidth,height=0.15\textwidth]{1296tr_grey_epoch220.pdf}
	%   \includegraphics[width=0.2\textwidth,height=0.15\textwidth]{1296tr_grey_epoch780.pdf}
	%}
	
	\subfigure[Global Channel Attention Map]{
		\includegraphics[width=0.8\textwidth]{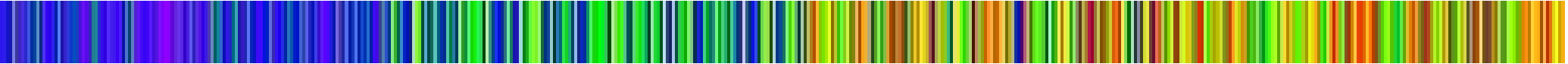}
	}
	\caption{(a) Column 1: one image with a red cross marked \textit{query pixel} which is a selected position in \textbf{\textit{Q}} shown in Fig.~\ref{fig:gscaPipline}. Column 2 to 5: related feature heatmaps computed with GSCA. Specifically, we use the corresponding vectors in \(\Phi\) and \(\Theta\) to compute attention maps according to Equation~\ref{con:intraAtten}. (b) Global Channel Attention Map displays the weight distribution along the channel.}
	\label{fig:GSCA_vis}
	%\vspace{-4pt}
\end{figure*}

To evaluate the effectiveness of the proposed NASK, we adopt two widely-used arbitrary-shaped text datasets and one multi-oriented text dataset for experiments. We also investigate NASK with detailed ablation studies.

\subsection{Dataset and Evaluation protocol}

\textbf{SynthText~\cite{gupta2016synthetic}} contains 800,000 natural images with the rendered text in various colors, fronts, scales and orientations. Generally, this dataset is used to pre-train the model.

\textbf{Total-Text~\cite{ch2017total}} is a comprehensive scene text dataset for  arbitrary-shaped texts. Except for the horizontal and multi-oriented texts, it contains large amount of curved texts. All images are annotated with word-level polygons and transcriptions. The training and testing sets are with 1255 and 300 images respectively. We use the updated official Python scripts\footnote{\url{https://github.com/cs-chan/Total-Text-Dataset/tree/master/Evaluation_Protocol}} to validate detection performance.

\textbf{SCUT-CTW1500~\cite{yuliang2017detecting}} is another widely benchmarked scene text dataset proposed in 2017. It consists of 1000 training images and 500 testing images. Compared to Total-Text, it involves both English and Chinese texts. The text instances from this dataset are annotated  with 14 boundary vertices. The evaluation script\footnote{\url{https://github.com/Yuliang-Liu/TIoU-metric/tree/master/curved-tiou}} is also provided by the official repository.

\textbf{ICDAR 2015 (IC15)~\cite{karatzas2015icdar}} is a commonly used dataset for multi-oriented text detection. It contains a total of 1500 pictures, 1000 of which are used for training and the remaining are for testing. The ground truth is annotated with word-level quadrangles. We also refer to the offcial online platform\footnote{\url{https://rrc.cvc.uab.es/?ch=4}}  for evaluation.

\begin{comment} 
For IC15, we utilize the standard evaluation protocols for multi-oriented text detection, namely \textit{precision} (\textit{P}), \textit{recall} (\textit{R}) and \textit{H-mean} (\textit{H}). Precisely, they are defined as follows:
\begin{equation}
P = \frac{TP}{TP+FP}
\end{equation}
\begin{equation}
R = \frac{TP}{TP+FN}
\end{equation}
\begin{equation}
H = 2\times \frac{P\times R}{P+R}
\end{equation}
where \(TP\),\(FP\),\(FN\) denotes the correctly classified text instances, incorrectly spotted instances, and the missing instances respectively. %Given a detected text instance \(b\), it will be assigned to the positive detection if the IOU between \(b\) and a ground truth is larger than a given threshold (normally set to 0.5). 
Since there is a trade-off between \(recall\) and \(precision\), $H$-$mean$ is a more objective measurement for performance assessments.    
\end{comment}
\subsection{Implementation Details}
\textbf{Network Structure.} For TIS, we choose the ImageNet~\cite{deng2009imagenet} pre-trained ResNet-50~\cite{he2016deep} as our backbone network with the last two down-sampling operations removed. For Geometry-aware RoI Alignment, we predefine the shape of the output feature map to be $8 \times 64$. The second segmentation network, namely FOX, is relatively simple with two up-sampling layers followed by one \(1\times1\) convolution with 6 output channels and the shape of the output feature map is $32 \times 256$.

\textbf{Training settings.}
%The proposed method is implemented in PyTorch~\cite{paszke2019pytorch}. 
%For all experiments, images are randomly cropped and resized into \(512\times512\). The cropped image regions are rotated randomly in 4 directions with \(0^\circ\), \(90^\circ\), \(180^\circ\), \(270^\circ\).
We  implement our method in PyTorch\footnote{\url{https://pytorch.org/}}. All experiments are conducted on four NVIDIA TitanX GPUs each with 12GB memory. The Adam optimizer is adopted here. We design a warm-up training strategy with the first segmentation network pre-trained on Synthetic dataset~\cite{gupta2016synthetic} for 10 epochs with learning rate set to $2 \times 10^{-4}$. This strategy leads to a precise first-stage segmentation, which is a prerequisite for the subsequent text shape refinement. Then the whole model including TIS, GeoAlign and FOX is fine-tuned with the initial learning rate $10^{-4}$ and the learning rate decay factor is set to 0.9.

For Total-Text, the training process is terminated after 10 epochs. We sample 8 center points in TCL and the group number of GSCA is set to 4. Thresholds $T_{tr}$, $T_{tcl}$ for regarding pixels to be text regions (for TIS) and TCL (for FOX) are set to (0.7, 0.6), respectively. 

The SCUT-CTW1500 dataset is also trained for 10 epochs. The TCL sampling points and  the GSCA group number are the same with those in Total-Text. $T_{tr}$, $T_{tcl}$ are found by grid search and are set to (0.8, 0.4). 

Since ICDAR 2015 dataset only contains the multi-oriented text instances, we reduce the TCL sampling points to 4 for simplification. The GSCA group number remains 4. $T_{tr}$, $T_{tcl}$ are set to (0.6, 0.5).

\subsection{Evaluation on Curved Text Benchmark}

We conduct experiments on Total-Text and SCUT-CTW1500  to verify the robustness of our method in detecting curved text. The quantitative results are shown in Table~\ref{table:qua_res}. 
  
On Total-Text dataset, NASK achieves impressive performance compared with state-of-the-arts. Specifically, it achieves the highest \textit{H}-\textit{mean} value (84.4\%) with \textit{FPS} reaching 8.4. Although LOMO achieves a higher \textit{precision} value than NASK, it has a fairly low \textit{recall} value, which means quite a few text instances are missed. Notably, compared with our conference version~\cite{cao2020all}, our method achieves 2.2\% performance gain in  \textit{H}-\textit{mean} with no reduction in efficiency. Besides, the quantitative results on SCUT-CTW1500 dataset also show NASK achieves a competitive result comparable to state-of-the-arts. Although the $recall$ and $precision$ value of NASK is inferior, it obtains the optimal $H$-$mean$. Since there is a trade-off between \(recall\) and \(precision\), $H$-$mean$ is a more objective measurement for performance assessments. For qualitative evaluation, some detection results are shown in Fig.~\ref{fig:exp_res}(a) and Fig.~\ref{fig:exp_res}(b).

\subsection{Evaluation on Multi-oriented Text Benchmark}

NASK is a general text detector and can be applied to the multi-oriented text benchmark as well. We verify the superiority of our method on the oriented text by conducting experiments on ICDAR 2015. %The number of TCL sampling points and the group number of GSCA are both set to 4. Threshold $T_{tr}$ and $T_{tcl}$ are set to (0.6, 0.5).
Quantitative and qualitative results are shown in Table~\ref{table:qua_res} and Fig.~\ref{fig:exp_res}(c), respectively. NASK achieves the $H$-$mean$ of 90.0\%, which consistently outperforms the previous state-of-the-art methods. Moreover, our method also obtains the best  $recall$ of 89.2\% among all methods in Table~\ref{table:qua_res}.

\subsection{Ablation studies}
In this section, we conduct several ablation studies to provide more insights about our design intuition. All the ablation experiments are performed on the SCUT-CTW1500 dataset. 

\subsubsection{Ablation study of GSCA} \qquad

\textbf{Effectiveness of GSCA.}
%We introduce the Group Spatial and Temporal Attention module (GSCA) into the network architecture to capture the long-range spatial and channel dependencies. 
We explore to reveal how GSCA helps. %For fair comparisons, we replace GSCA with two stacked $1\times1$ convolution layers so that they share almost the same computation overhead. 
Specifically, we apply a set of comparative experiments with different $G$ values. As shown in Table~\ref{table:ablation} (a), GSCAs with all $G$ values lead to the performance gain in \textit{H-mean} compared to the native model (list in Table~\ref{table:ablation} (c) with \textit{Attention} set to None). For instance, by setting $G$ to 4, $H$-$mean$ improves by 2.2\%. In this case, it strikes a balance between performance and efficiency. Therefore, the group attention mechanism enhances the long-range relationship and boosts the detection accuracy with the affordable overhead. 
%the model without the attention module leads to a obvious drop in \textit{H-mean} compared with the method with GSCA. %

We also present the visualization results of GSCA. In Fig.~\ref{fig:GSCA_vis}(a), we visualize the group-wise spatial attention map. Specifically, we randomly select one pixel in the input image and regard it as the \textit{query pixel}  in the feature map \textbf{\textit{Q}} shown in Fig.~\ref{fig:gscaPipline}.  Then we use the corresponding vectors in $\Phi$ and $\Theta$ to compute attention maps according to Equation~\ref{con:intraAtten}. The results indicate that GSCA is context-aware \textit{i.e.,} most of the weights are focused on the pixels belonging to the same category with \textit{query pixel}. Fig.~\ref{fig:GSCA_vis}(b) presents the channel-wise weight distribution computed by the Global Channel Attention branch, which helps the training process focus on the most discriminative channels. 

\begin{comment}
\begin{table}[!ht]
    \caption{Ablation studies on SCUT-CTW 1500}\label{tab:tablenotes}
    \centering
    \begin{threeparttable}
     
\begin{tabular}{|c|c|c|c|c|c|c|}
\hline
index & \(1^{st} seg\) & \textit{\(G\)} & \textit{R} & \textit{P} & \textit{H} & \textit{F} \\
\hline
\multirow{6}*{(a)} & \multirow{6}*{\ding{51}} & 0 & 76.4 &  79.7 & 78.0 & 16.3 \\
~ & ~ & 2 & 79.3 & 83.2 & 81.2 & 3.4 \\
~ & ~ & 4 & 78.3 & 82.8 & 80.5 & 12.1 \\
~ & ~ & 8 & 78.1 & 82.3 & 80.1 & 12.5 \\
~ & ~ & 12 & 77.9 & 82.8 & 80.3 & 12.7 \\
~ & ~ & 16 & 77.3 & 81.6 & 79.4 & 12.9 \\
\hline
\multirow{2}*{(b)} & \ding{55} & 4 & 72.7 & 77.5 & 75.0 & 16.3 \\
~ & \ding{51} & 4 & 78.3 & 82.8 & 80.5 & 12.1 \\
\hline 
\end{tabular} 
      \begin{tablenotes}
        \footnotesize
        \item Note: \(1^{st} seg\) means the first stage segmentation network namely \textit{TIS}; \(G\) denotes the group number of the attention module. %\textit{R},\textit{P},\textit{H},\textit{F} is the same as Table 1.
        %\item[2] .
      \end{tablenotes}
    \end{threeparttable}
\end{table}
\end{comment}

\textbf{Influence of the number of attention module groups \textit{G}.} There is a trade-off between the accuracy and the speed when setting the different group numbers of GSCA. Intuitively, less grouping leads to higher accuracy and lower speed, and vice versa. %However, the results list in Table 4 show that the truth is not always as imagined. Actually, the detection speed increases when the group number increases as expected. 
In Table~\ref{table:ablation} (a), as expected, the detection speed increases with the rise of the group number and reaches the limit at about 13.8 \textit{FPS}. Notably, it is noticed that the quantitative performance is not much sensitive to $G$ when $G \geq 4$. This may be explained by the fact that the Global Channel Attention branch in Fig.~\ref{fig:gscaPipline} effectively captures the rich correlations among groups, thus alleviating the negative effects of the grouping operation. %Besides, when the feature map is split into more groups, it facilitates the restricted optimization and eases the training.

\textbf{GSCA vs. DANet vs. CCNet.} To demonstrate the superiority of GSCA, we replaced GSCA with two widely used self-attention modules, DANet~\cite{fu2019dual} and CCNet~\cite{huang2019ccnet}, respectively, while keeping the rest of the network unchanged. The comparison results list in Table~\ref{table:ablation} (c) show that the model equipped with GSCA outperforms that with CCNet, which may be due to the fact that CCNet only considers the spatial correlations. While DANet shares the similar $H$-$mean$ with our GSCA, it has a lower \textit{FPS} than GSCA (10.3 \textit{vs.} 12.1). Therefore, in our task, GSCA is a more effective and efficient attention module compared with the state-of-the-arts. Besides, we also compare GSCA to our conference version $\text{GSCA}_{conf}$ and the results show that GSCA outperforms $\text{GSCA}_{conf}$ with no degrade in speed. 

\subsubsection{Ablation study of Geometry-aware RoI Alignment}  \qquad \qquad
\textbf{Influence of GeoAlign.}
The proposed Geometry-aware RoI Alignment module effectively selects the sampling points and avoids the background interference. We conduct comparison experiments by replacing our Geometry-aware RoI Pooling with RoI Align~\cite{he2017mask} and RoI Pooling~\cite{ren2015faster}, respectively. The results list in Table~\ref{table:ablation} (d) demonstrate that the GeoAlign-equipped model achieves the best performance (1.6\% $H$-$mean$ gain compared to RoI Pooling) among the three variants with negligible overheads. Therefore, GeoAlign generates more informative features for the following geometric attribute prediction.

\subsubsection{Ablation study of Fiducial pOint eXpression module}
\begin{figure}[htb]
 \centering
  %\centerline{\includegraphics[width=6cm]{experiments/ablation/num.pdf}}
  \centerline{\includegraphics[width=0.9\linewidth]{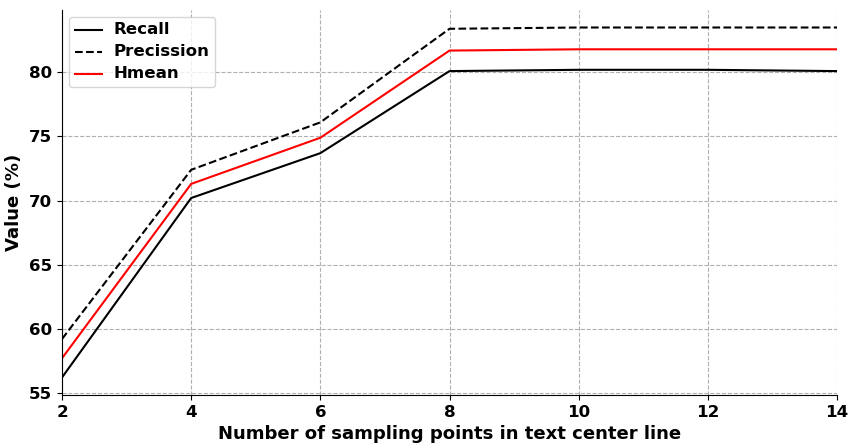}}
 % with channel attention and result
%   \centerline{\includegraphics[width=9cm]{attentionGroupWithChannel.pdf}}
  %\centerline{attentionGroup}\medskip
\caption{Ablation studies of the number of sample points.}
\label{fig:ab_n}
\end{figure}

\textbf{Influence of the number of sample points \textit{n}.}
With our Fiducial Point Expression module, the curve text representation is decided by a set of $2n$ fiducial points, thus the number of text center line sample points is an important hyper-parameter. To explore this, we evaluate the performance under different values of $n$. The results shown in Fig.~\ref{fig:ab_n} witness a sustained increase when $n$ changes from 2 to 8 and then the performance gradually converges. Therefore, we set $n$ to 8 in our experiments.

\subsubsection{Ablation study of the two-stage architecture design} \qquad  
\textbf{Effectiveness of the two-stage segmentation.} 
To demonstrate the efficacy of our two-stage segmentation architecture, we conduct comparison experiments that only apply the first or the second segmentation network. (1) When only applying the first stage segmentation, it falls into a simple segmentation task. Specifically, we additionally append a convolution layer with 2 output layers (for text and non-text prediction) and a Sigmoid activation function. Then we threshold the output feature map and use \textit{approxPolyDP} in OpenCV to obtain the bounding box. (2) For the experiment with only the second-stage segmentation network, we directly apply FOX on the input image and reconstruct the curved text instances.

The comparison results in Table~\ref{table:ablation} (b) show that the performance of the two-stage segmentation surpasses the single-stage segmentation by a large margin. The varient with only the first stage segmentation can not describe arbitrary-shaped text instances accurately, thus having inferior performance. For the other varient which drops the first stage of rectangle text instance segmentation, the geometric properties FOX refers to need to be predicted on the input image directly. Compared with NASK which conducts the segmentation on the RoI feature maps, this method introduces more background interference and leads to the decrease of detection accuracy (75.8\% \textit{vs.} 81.7\% in $H$-$mean$). Based on the above analysis, our two-stage segmentation architecture effectively decompose the non-trival arbitrary-shaped text detection into two stages, namely rectangle text proposal detection and text refinements. This coarse-to-fine manner leads to better performance compared to applying one of the two stages alone.

\begin{comment} 
\begin{figure}[htb]
 \centering
  \centerline{\includegraphics[width=8cm]{experiments/ablation/vsTextSnake.pdf}}
  %\centerline{(a) vsTextSnake}\medskip
\caption{Fiducial point generation result using the proposed method (Up) and the detection result of TextSnake (Down).}
\label{fig:res}
\end{figure}   
\end{comment}

\section{Conclusion}
In this paper, we propose a novel two-stage segmentation-based text detector NASK to facilitate arbitrary-shaped text detection. We firstly leverage a text instance segmentation network TIS to obtain the rectangle proposals. To capture the long-range dependency, a self-attention based mechanism called Group Spatial and Channel Attention module (GSCA) is incorporated into TIS to augment the feature representation. Then Geometry-aware Text RoI Alignment (GeoAlign), a reformative alternative for RoI Align, is applied to warp the rectangle text proposals to the fixed size. Finally, we propose a Fiducial Point Expression module (FOX) which utilizes fiducial points to represent the arbitrary-shaped texts. Experiment results on both the multi-oriented and curved text datasets have demonstrated the effectiveness and efficiency of our proposed NASK.

\bibliography{ref.bib}

% or


\begin{comment}\\

\appendices
\section{Proof of the First Zonklar Equation}
Appendix one text goes here.

\section{}
Appendix two text goes here.
\end{comment}

\section*{Acknowledgment}
This paper was partially supported by the IER foundation (No. HT-JD-CXY-201904) and  Shenzhen Municipal Development and Reform Commission (Disciplinary Development Program for Data Science and Intelligent Computing). Special acknowledgements are given to Aoto-PKUSZ Joint Lab for its support.

\ifCLASSOPTIONcaptionsoff
  \newpage
\fi

\end{document}